\documentclass[journal]{IEEEtran}
\usepackage{amsmath,amsfonts,amssymb}
\usepackage[linesnumbered,ruled]{algorithm2e}
\usepackage{amsmath}
\usepackage{array}
\usepackage[caption=false,font=normalsize,labelfont=sf,textfont=sf]{subfig}
\usepackage{textcomp}
\usepackage{stfloats}
\usepackage{url}
\usepackage{verbatim}
\usepackage{graphicx}
\usepackage{bm}
\usepackage{cite}
\usepackage{multirow}
\usepackage{color}
\usepackage{mathtools}
\usepackage{amsthm}
\newtheorem*{theorem}{Theorem}

\newtheorem{remark}{Remark}

\DeclareMathOperator*{\argmax}{argmax}
\DeclareMathOperator*{\argmin}{argmin}
\newcommand\sbullet[1][.5]{\mathbin{\vcenter{\hbox{\scalebox{#1}{$\bullet$}}}}}

\begin{document}

\title{($\boldsymbol{\theta}_l, \boldsymbol{\theta}_u$)-Parametric Multi-Task Optimization: Joint Search in Solution and Infinite Task Spaces}


\author{Tingyang Wei,
        Jiao Liu,
        Abhishek Gupta,\IEEEmembership{~Senior Member,~IEEE},
        Puay Siew Tan,
        and~Yew-Soon Ong,\IEEEmembership{~Fellow,~IEEE}
\thanks{This research is partly supported by the Distributed Smart Value Chain programme under the Singapore RIE2025 Manufacturing, Trade and Connectivity (MTC) Industry Alignment Fund-Pre-Positioning (Award No: M23L4a0001), the MTI under its AI Centre of Excellence for Manufacturing (AIMfg) (Award W25MCMF014), the National Research Foundation, Singapore and DSO National Laboratories under the AI Singapore Programme (AISG Award No.: AISG2-GC-2023-010, ”Design Beyond What You Know”: Material-Informed Differential Generative AI (MIDGAI) for Light-Weight High-Entropy Alloys and Multi-functional Composites (Stage 1b)”, the Centre for Frontier AI Research (CFAR) under Agency for Science, Technology and Research (A*STAR), and the College of Computing and Data Science, Nanyang Technological University. Furthermore, this work is also supported in part by the Ramanujan Fellowship from the Anusandhan National Research Foundation, Government of India (Grant No. RJF/2022/000115). (\textit{Corresponding author: Abhishek Gupta})}
\thanks{T. Wei and J. Liu are with the College of Computing and Data Science, Nanyang Technological University, Singapore (e-mail: TINGYANG001@e.ntu.edu.sg, jiao.liu@ntu.edu.sg)}
\thanks{A. Gupta is with the School of Mechanical Sciences, Indian Institute of Technology, Goa, India (e-mail: abhishekgupta@iitgoa.ac.in)}
\thanks{P. S. Tan is with the Singapore Institute of Manufacturing
Technology (SIMTech), Agency for Science, Technology and
Research, Singapore (e-mail: pstan@simtech.a-star.edu.sg)}
\thanks{Y.-S. Ong is with the College of Computing and Data Science, Nanyang Technological University, Singapore, and also with the Centre for Frontier AI Research (CFAR), Agency for Science, Technology and
Research, Singapore (e-mail: asysong@ntu.edu.sg)}}
\markboth{Journal of \LaTeX\ Class Files,~Vol.~14, No.~8, August~2021}%
{Shell \MakeLowercase{\textit{et al.}}: A Sample Article Using IEEEtran.cls for IEEE Journals}

\maketitle
\begin{abstract}
Multi-task optimization is typically characterized by a fixed and finite set of tasks.
The present paper relaxes this condition by considering a non-fixed and potentially infinite set of optimization tasks defined in a parameterized, continuous and bounded task space. 
We refer to this unique problem setting as parametric multi-task optimization (PMTO).
Assuming the bounds of the task parameters to be ($\boldsymbol{\theta}_l$, $\boldsymbol{\theta}_u$), a novel ($\boldsymbol{\theta}_l$, $\boldsymbol{\theta}_u$)-PMTO algorithm is crafted to operate in two complementary modes.
In an offline optimization mode, a joint search over solution and task spaces is carried out with the creation of two approximation models: (1) for mapping points in a unified solution space to the objective spaces of all tasks, which provably accelerates convergence by acting as a conduit for inter-task knowledge transfers, and (2) for probabilistically mapping tasks to their corresponding solutions, which facilitates evolutionary exploration of under-explored regions of the task space.
In the online mode, the derived models enable direct optimization of any task within the bounds without the need to search from scratch.
This outcome is validated on both synthetic test problems and practical case studies, with the  significant real-world applicability of PMTO shown towards fast reconfiguration of robot controllers under changing task conditions.
The potential of PMTO to vastly speedup the search for solutions to minimax optimization problems is also demonstrated through an example in robust engineering design.
\end{abstract}


\section{Introduction}

\IEEEPARstart{O}{ptimization} tasks rarely exist in isolation. Multi-task optimization (MTO), first formulated in evolutionary computation by means of the multifactorial evolutionary algorithm \cite{mfea}, has therefore emerged as a promising approach for solving problems simultaneously by leveraging shared information across related tasks \cite{ong-discuss}. 
This concept has been successfully applied in both evolutionary computation \cite{mfea} and Bayesian optimization \cite{MTBO}, supported by knowledge transfer strategies such as solution-based transfer~\cite{moomfea, mfea2} and model-based transfer~\cite{MTBO, lda, autoencoding}. 
The applicability of these methods has been studied in a variety of domains, including robotics \cite{osaba-rl}, time series prediction \cite{mfgp}, vehicle shape design \cite{car_new}, among many others \cite{dozen}, achieving faster convergence under limited computational budgets.

\begin{figure*}[!t]
\centering
\subfloat[Multi-task Optimization]{\includegraphics[width=2.40in]{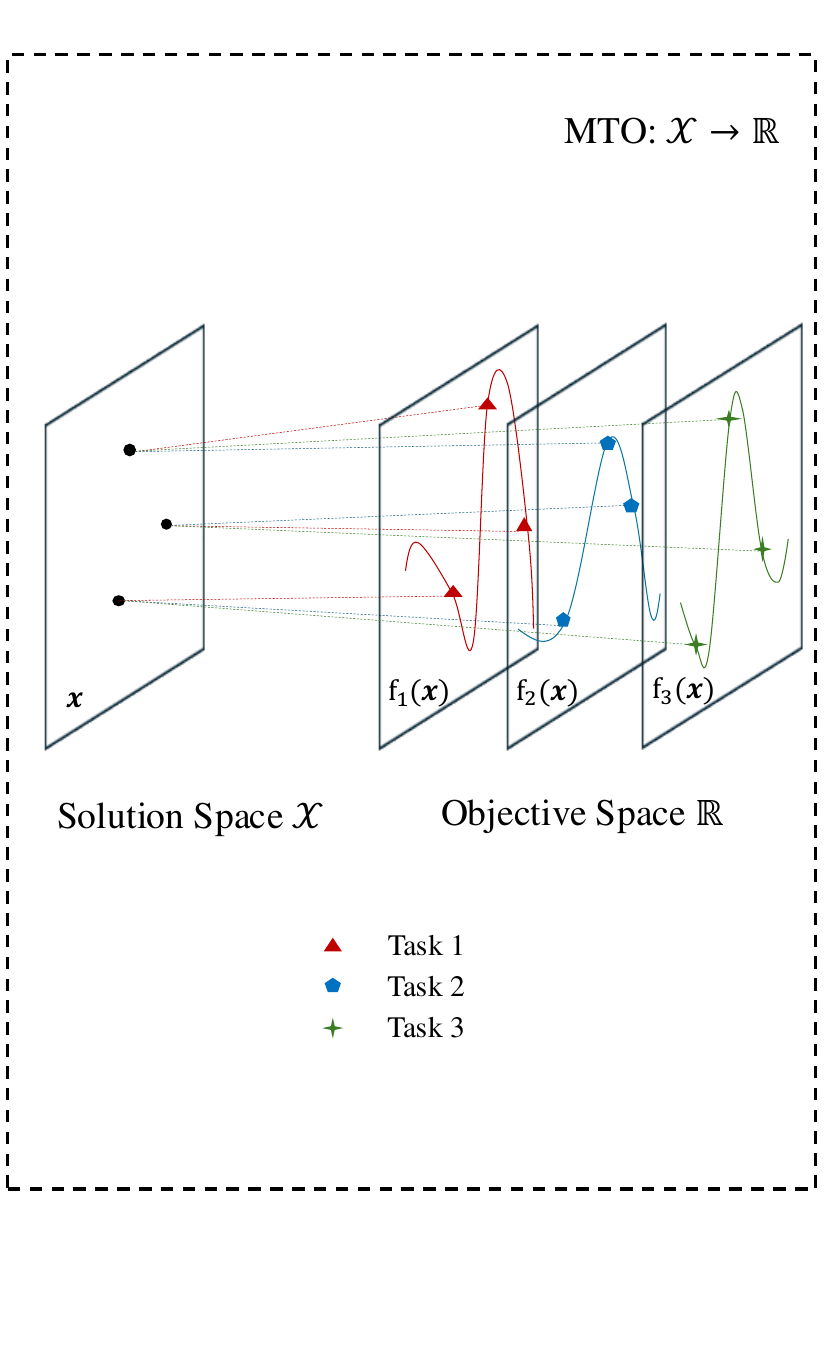}%
\label{fig: MTO}}
\hfil
\subfloat[Parametric Multi-task Optimization]{\includegraphics[width=3.60in]{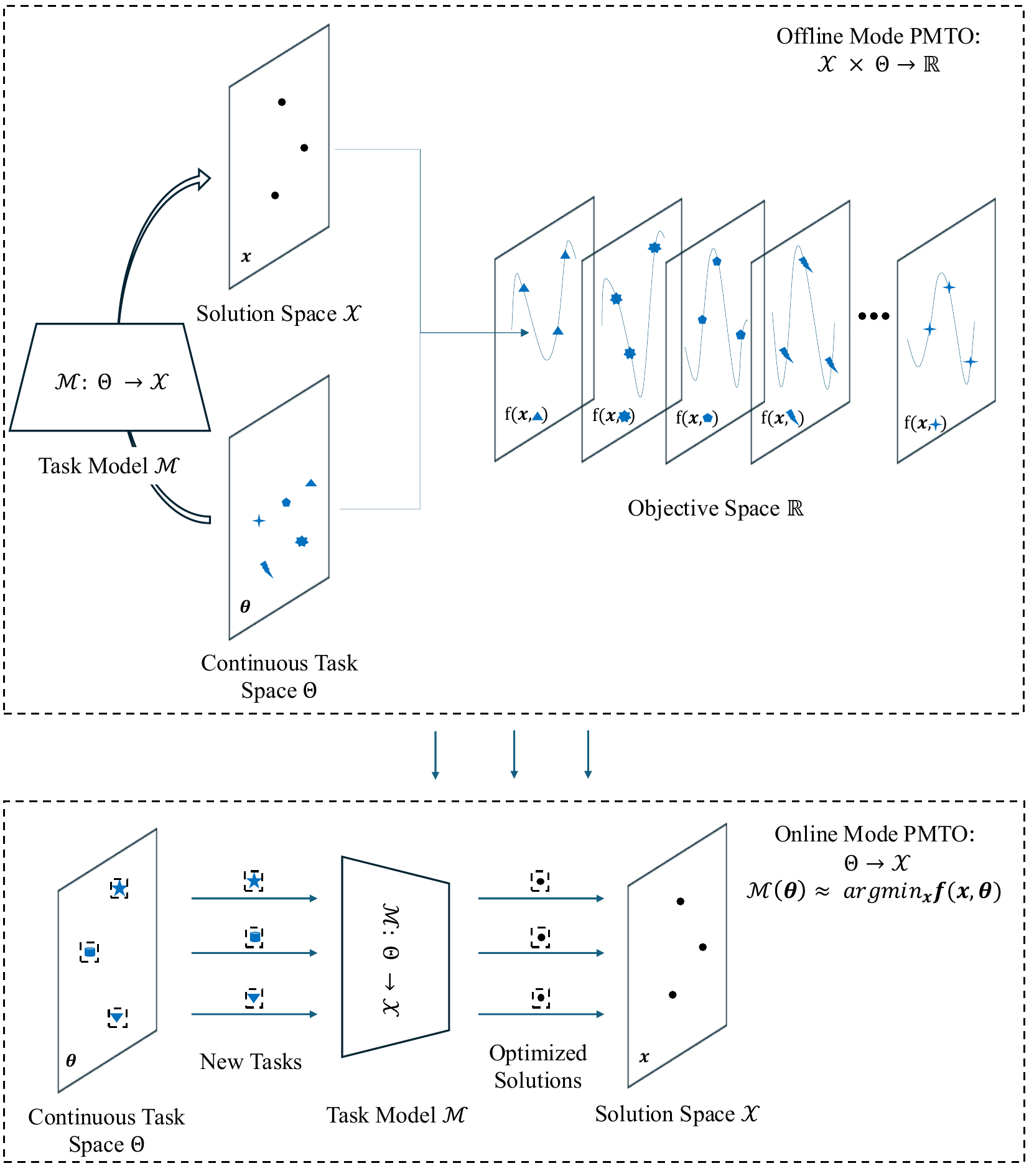}%
\label{fig: PMTO}}
\caption{Distinguishing multi-task optimization and parametric multi-task optimization. (a) In multi-task optimization, there typically exists a predefined and fixed set of
tasks (3 in this example) whose objective functions map points from a unified solution space to the respective objective spaces. (b) In parametric multi-task optimization, continuous task parameters imply a potentially infinite set of tasks. 
During PMTO's offline optimization mode, a task model $\mathcal{M}: \Theta \rightarrow \mathcal{X}$ is built to actively evolve and optimize tasks with unknown optima. 
In the online mode, this task model is applied to directly predict optimized solutions for any new task without entailing additional evaluation costs.}
\label{fig: example}
\end{figure*}

MTO is typically characterized by a fixed and finite set of optimization tasks. 
In this paper, we relax this paradigm by introducing \textit{Parametric Multi-Task Optimization~(PMTO)} as a framework that considers a non-fixed and potentially infinite set of optimization tasks within a parameterized, continuous, and bounded task space $\Theta \subset \mathbb{R}^D$.
Each task in PMTO is defined by a unique vector $\boldsymbol{\theta} \in (\boldsymbol{\theta}_l, \boldsymbol{\theta}_u)$, where $\boldsymbol{\theta}$ represents the \textit{task parameters}. 
Fig.~\ref{fig: example} illustrates the difference between MTO and PMTO. 
As shown in Fig.~1(a), the objective functions in MTO are defined as independent mappings from a unified decision space to task-specific objective spaces.
In contrast, the task parametrization directly influences the mapping from solutions to objectives in PMTO. 
This inclusion of a task space brings opportunities and challenges.
PMTO benefits from leveraging information from both solution and task representations, enabling more effective inter-task relationship analysis and potentially more efficient cross-task optimization. 
However, continuous task parameters also imply an infinite task set, posing a challenge for algorithm development compared to the simpler case of a fixed task set in MTO.

To address PMTO problems, this paper introduces a novel ($\boldsymbol{\theta}_l, \boldsymbol{\theta}_u$)-PMTO algorithm that incorporates a joint search over solutions and tasks.
The algorithm operates in two complementary modes. In an \emph{offline} optimization mode, a mapping $f: \mathcal{X} \times \Theta \rightarrow \mathbb{R}$ from the solution space to the objective space is built by including task parameters as side information, thereby capturing correlations and facilitating the transfer of knowledge across tasks for provably faster convergence.
($\boldsymbol{\theta}_l, \boldsymbol{\theta}_u$)-PMTO also induces a probabilistic model $\mathcal{M}: \Theta \rightarrow \mathcal{X}$ mapping tasks to their corresponding optimized solutions, with the inter-task relationships captured by the model helping support the evolutionary exploration of undersampled tasks. 
Through an iterative search process, the algorithm collects data on representative points that cover the task space. 
A well-calibrated task model $\mathcal{M}$ obtained by the end of the offline ($\boldsymbol{\theta}_l, \boldsymbol{\theta}_u$)-PMTO run can then enable fast online identification of near-optimal solutions for any task in that space. 

The practical implications of PMTO are particularly profound in its online application to problems requiring fast adaptation in diverse and uncertain environments.
Taking robotic control as an example~\cite{qd-multitask, nature}, standard MTO can be configured for multiple predefined morphologies of robots to achieve specific tasks including navigation, manipulation, and locomotion.
However, real-world scenarios often involve unexpected changes, such as unforeseen damages or obstacles.
These changes may cause the actual problem setup to deviate significantly from the predefined MTO set, leading to inferior results when obtained solutions are applied to unforeseen tasks.
PMTO addresses this challenge seamlessly by pre-optimizing over a non-fixed and potentially infinite set of tasks, as shown in the top panel in Fig.~1(b).
This enables diverse operating morphologies or damage conditions to be encoded as continuous task parameters, allowing the resulting task model to directly predict optimized solutions for any unseen problem setting, as illustrated in the bottom panel in Fig.~1(b).

PMTO seeks to simultaneously explore solutions for a range of parameterized tasks, making it a versatile framework for addressing problem classes where such task spaces naturally occur.
Consider minimax optimization in robust engineering design~\cite{zhou_saea} as an example, where the goal is to identify designs that are worst-case optimal under random variations in the design parameters.
This involves optimizing (minimizing) the worst (maximum) objective value to account for real-world variabilities.
By re-imagining the space of random variations as the solution space ($\mathcal{X}$) and the original design parameters to form the task space ($\Theta$), the robust optimization problem can be recast in the new light of PMTO.
Related mathematical programs, such as bilevel programming which is known to be amenable to MTO approaches~\cite{bilevel}, can also be recast in this way. 
A case study demonstrating this novel application to minimax optimization is presented in Section VI-E.

In this paper, we focus on expensive optimization problems in PMTO settings.
Assessing the performance of derivative-free optimizers under stringent budget constraints is deemed especially meaningful, as real-world problems frequently entail costly function evaluation calls.
The major contributions of this paper are summarized below.
\begin{itemize}
    \item We introduce the concept of PMTO as a generalization of MTO. The distinctive feature of this paradigm is the inclusion of a continuous task space, resulting in a potentially infinite number of optimization tasks within that space.
    \item We theoretically and empirically show that incorporating continuous task parameters into the optimization loop enables faster convergence even when optimizing a fixed set of tasks, compared to optimizing them separately and independently.
    \item We propose a ($\boldsymbol{\theta}_l, \boldsymbol{\theta}_u$)-PMTO algorithm by coupling multi-task optimization in the solution space with a strategic task evolution component guided by an iteratively updated task model. The algorithm facilitates data acquisition and training of a well-calibrated task model~($\mathcal{M}$) capable of predicting optimized solutions for any task parameterized by $\boldsymbol{\theta} \in (\boldsymbol{\theta}_l, \boldsymbol{\theta}_u)$. 
    \item Rigorous experimental investigation of the overall method is carried out on synthetic test problems with varied properties. The utility of the derived task model is showcased in different real-world applications spanning adaptive control systems and robust engineering design.
\end{itemize}

The remainder of this paper is organized as follows. 
Section II reviews related works on multi-task optimization and parametric programming.
Section III presents preliminaries including standard multi-task optimization, the parametric multi-task optimization formulation, and approximations via probabilistic Gaussian process (GP) models.
Section IV discusses a restricted version of PMTO with a fixed and finite set of parameterized tasks, theoretically proving that the inclusion of task parameters accelerates convergence.
Section V introduces our novel algorithm, dubbed ($\boldsymbol{\theta}_l, \boldsymbol{\theta}_u$)-PMTO, for jointly searching over both solution and task spaces.
Section VI empirically verifies the effectiveness of the method on a variety of synthetic and practical problems, establishing PMTO as a promising direction for future research. 
Section VII concludes the paper.

\section{Related Works}\label{sec: related}
\subsection{Multi-task Optimization}
MTO deals with solving a fixed set of tasks simultaneously within limited evaluation budgets. 
In the Bayesian optimization literature, MTO usually leverages GPs with multi-task kernels~\cite{MTBO, moo-finv}, modeling both data inputs and task indices to capture inter-task relationships.
A multi-task GP~\cite{MTGP} is equipped to transfer information from related source tasks to enhance model performance in a target task, with associated multi-task Bayesian optimization algorithms demonstrating their effectiveness in both unconstrained~\cite{confidencebo} and constrained optimization settings~\cite{safebo}.

In evolutionary computation, multi-task optimization~\cite{mfea, wei} has proven to be effective for high-dimensional~\cite{ong-discuss, multi-benchmark, mfea2, momfea2} and combinatorial optimization problems~\cite{multiform, yuxiao, combinatorial-multitask}.
Leveraging the implicit parallelism of evolutionary algorithms~\cite{mfea, ong-discuss}, evolutionary multi-task optimization facilitates knowledge transfer by either implicitly sharing high-quality solutions across optimization tasks~\cite{mfea, moomfea} or explicitly mapping solutions from source domains to target domains~\cite{lda, autoencoding}. 
In solution-based transfer, the solution distribution of a source task may be shifted according to handcrafted translation vectors~\cite{gmfea, fuzzy} or the multi-task search may be guided by the maximum point of the product of population distributions of a source-target task pair~\cite{liang-1}.
In model-based knowledge transfer, a pair-wise mapping from a source to the target task is established by distinct learning-based methods including least square methods~\cite{lda, autoencoding}, subspace alignment~\cite{gong2}, manifold alignment~\cite{sysu}, geodesic flow~\cite{geodesic-flow}, among others~\cite{da2, domain}.
Building on these knowledge transfer strategies, evolutionary multi-task optimization has shown promise in a plethora of applications \cite{dozen}, including job shop scheduling~\cite{zhixing}, sparse reconstruction~\cite{reconstruction1}, point cloud registration~\cite{wuyue}, and recommender systems~\cite{recommendation}. 

Despite considerable research efforts, existing methods primarily focus on traditional MTO problems with a fixed and finite set of optimization tasks, without considering parameterizations of the tasks themselves.
Recent studies have made initial progress toward filling this gap.
The concept of modeling a parameterized family of tasks in sequential transfer optimization was discussed in~\cite{xue-family}.
Parametric-task MAP-Elites\cite{multitask-qd} seeks to illuminate a continuous task space through an exhaustive search over representative tasks that cover the space, but the cumulative cost of the method under expensive evaluations makes it impractical for many applications.
Multi-scenario optimization~\cite{multitask-robust} aims to collaboratively solve an iteratively updated set of scenarios via multifactorial evolutionary algorithms~\cite{mfea}. 
Yet, it fails to leverage task-specific side information and cannot provide solutions for new tasks beyond those encountered during optimization. In contrast, the proposed PMTO framework not only enables data-efficient joint search over continuous solution and task spaces but also produces a task model that enables fast solution prediction for new tasks under tight evaluation budgets.

\subsection{Parametric Programming}
Parametric programming is a type of mathematical optimization framework that seeks to derive optimal solutions as a function of uncertain (task) parameters~\cite{mp-survey}.
Associated techniques have been applied in diverse optimization contexts including linear programming~\cite{mp-lp}, mixed integer programming~\cite{mp-mip}, or nonlinear programming~\cite{mp-nlp, mp-survey}.  
Generally coupled with model predictive control frameworks~\cite{mp-mpc, RAL-CBO}, parametric programming has shown promising results in adaptive control systems encompassing energy management of hybrid vehicles~\cite{energy-manage-vehicle}, autonomous steering control~\cite{autonomous-steering-control}, current control in power electronics~\cite{current-control}, and endoscope control in biomedical cases~\cite{magnetic-endoscope}, among others~\cite{mp-survey}.
Being an active but nascent area of research, parametric programming techniques typically limit to problems that possess a precise mathematical description, thus precluding application to problems that involve expensive, nonlinear and non-differentiable objective functions.
Our proposed ($\boldsymbol{\theta}_l, \boldsymbol{\theta}_u$)-PMTO algorithm addresses this gap by inducing computationally cheap approximations in a joint exploration of solution and parametrized task spaces, thereby distinguishing from any related work in the literature.

\subsection{Contextual Bayesian Optimization}
Contextual Bayesian Optimization (CBO) extends contextual bandit frameworks~\cite{nips-ong, kernel-bandits} to contextual black-box optimization via probabilistic surrogate models~\cite{offline-contextual, meta-cbo}.
Closely related representative works include offline CBO~\cite{offline-contextual} and continuous multi-task Bayesian optimization~\cite{kg-base, aq-kg}. 
In contrast to this line of research, PMTO is designed to address infinitely many optimization tasks through a unified approach featuring dual working modes. 
The method assumes all tasks lying within certain prespecified bounds to be equally likely to occur.
During offline optimization over this task space, PMTO performs joint exploration over solutions and tasks, guided by an iteratively updated task model to enhance search efficiency and task coverage. 
In online mode, the derived task model $\mathcal{M}: \Theta \rightarrow \mathcal{X}$ provides immediate predictions of optimized solutions for any new incoming task, eliminating the need for additional costly evaluations. 
This dual offline–online design enables PMTO to flexibly tackle continuous task scenarios while maintaining evaluation efficiency.

\section{Preliminaries}
\subsection{Multi-task Optimization}
Standard MTO aims to simultaneously solve $M$ optimization tasks, $\mathcal{T}_1, \mathcal{T}_2, \ldots, \mathcal{T}_M$. 
The group of tasks can be formulated as follows:
\begin{equation}
    \argmin_{\mathbf{x}_m \in \mathcal{X}_m} f_m(\mathbf{x}_m),~~m \in [M],
\label{eq: traditional-mto}
\end{equation}
where the decision vector $\mathbf{x}_m$ of the $m$-th optimization task lies in the search space $\mathcal{X}_m$ and optimizes the objective function $f_m: \mathcal{X}_m \rightarrow \mathbb{R}$~\cite{mfea}.
It is commonly assumed in MTO that no prior knowledge or side information about the tasks is available. 
As a result, evolutionary or Bayesian MTO algorithms must adaptively identify task similarities or correlations during the optimization process to enhance convergence and mitigate the risk of negative transfer~\cite{ong-discuss, mfea2}.

\subsection{Parametric Multi-task Optimization}
PMTO contains additional task parameters as side information, leading to the following formulation:
\begin{equation}
    \argmin_{\mathbf{x} \in \mathcal{X}} f(\mathbf{x}, \mathbf{\boldsymbol{\theta}}),~~\forall \boldsymbol{\theta} \in \Theta,
\label{eq: parameterized-mto}
\end{equation}
where $\Theta \subset \mathbb{R}^D$ is a bounded continuous space such that each optimization task is represented by a unique parameter vector $\boldsymbol{\theta} \in \Theta$.
Comparing (\ref{eq: parameterized-mto}) to the formulation (\ref{eq: traditional-mto}), two major differences between MTO and PMTO surface.
\begin{itemize}
    \item The MTO problem does not assume any prior knowledge or side information about the constitutive optimization tasks. In contrast, PMTO assumes side information in the form of task parameters to be available and usable in the optimization loop. 
    \item MTO limits the search process within the solution spaces $\{\mathcal{X}_m\}_{m=1}^{M}$ of a fixed set of optimization tasks $\{\mathcal{T}_m\}_{m=1}^{M}$. PMTO generalizes this idea by considering a task space containing potentially infinite tasks characterized by continuous task parameters, i.e., $\forall \boldsymbol{\theta} \in \Theta$.
\end{itemize}

In PMTO, we propose to address the problem defined in (\ref{eq: parameterized-mto}) by actively solving a representative set of tasks in $\Theta$ during an offline optimization mode.
Through this process, a task model $\mathcal{M}: \Theta \rightarrow \mathcal{X}$ is established to map task parameters to their corresponding optimized solutions. 
Once constructed, this model operates in the online mode, where it directly predicts solutions for any incoming task parameter in 
$\Theta$ without the need for additional costly evaluations. 
Specifically, the task model provides approximations to solutions of (\ref{eq: parameterized-mto}) as:
\begin{equation}
    \forall \boldsymbol{\theta} \in \Theta,~~\mathcal{M}(\boldsymbol{\theta}) \approx \mathbf{x}_{\boldsymbol{\theta}}^* \coloneqq  \argmin_{\mathbf{x} \in \mathcal{X}} f(\mathbf{x}, \boldsymbol{\theta}).
\label{eq: task_model}
\end{equation}
It is important to note that the task model ($\mathcal{M}: \Theta \to \mathcal{X}$) is explicitly trained to approximate the mapping from each task parameter $\boldsymbol{\theta}$ to its corresponding optimized solution $ \mathbf{x}_\theta^\star \coloneqq \arg\min_{\mathbf{x} \in \mathcal{X}} f(\mathbf{x}, \boldsymbol{\theta})$, instead of inferring \( \mathbf{x}^\star \) via optimization over the surrogate models that approximate the ground truth $f$.
The task model can then be directly utilized online to predict solutions for new task parameters. 
This capability allows PMTO to provide rapid responses to new tasks with limited computational overhead.

\begin{algorithm}[!t]
\label{alg: GP-UCB}
\DontPrintSemicolon
    \caption{GP-based Optimization}
    \KwData{Initialization budgets $N_{init}$, Total evaluation budgets $N_{tot}$, Objective function $f$;}
    \KwResult{The best solution found in solution set $\mathcal{D}$;}
    Randomly initialize $N_{init}$ solutions $\{ \textbf{x}^{(i)} \}_{i=1}^{N_{init}}$\;  
    Evaluate initial solutions via the objective function $f(\sbullet)$\;
    $\mathcal{D} \gets \{(\mathbf{x}^{(i)}, f(\mathbf{x}^{(i)}))\}_{i=1}^{N_{init}}$ \;
    $t \gets 0$\;
    Train the GP model on $\mathcal{D}$\;
    \While{$t + N_{init}<N_{tot}$}
    {
    Sample new input data $\mathbf{x}^{(t)}$ based on (\ref{eq: acquistion})\;
     $\mathcal{D} \gets \mathcal{D} \cup \{(\mathbf{x}^{(t)}, f(\mathbf{x}^{(t)}))\}$ \;
    Update the GP model according to $\mathcal{D}$\;
    $t \gets t + 1$\;
    }

\end{algorithm}
\subsection{Gaussian Process}
Throughout the paper, GP models~\cite{GPML} are adopted for principled probabilistic function approximation.
Given an unknown function $f$, the GP assumes $f$ to be a sample of a Gaussian prior, i.e., $f \sim \mathcal{GP}(\mu(\sbullet), \kappa(\sbullet, \sbullet))$, defined by the mean function $\mu(\mathbf{x}) = \mathbb{E}[f(\mathbf{x})]$ and the covariance function $\kappa(\mathbf{x}, \mathbf{x}') = Cov[f(\mathbf{x}), f(\mathbf{x}')]$.
Given a dataset $\mathcal{D} = \{(\mathbf{x}^{(i)}, y^{(i)})\}_{i=1}^{N}$, where $y^{(i)} = f(\mathbf{x}^{(i)}) + \epsilon^{(i)}$ and $\epsilon^{(i)} \sim \mathcal{N}(0, \sigma_{\epsilon}^2)$, the predicted posterior distribution of the GP, i.e., $\mathcal{N}(\mu({\mathbf{x}}),\sigma^2({\mathbf{x}}))$, at an unseen query point ${\mathbf{x}}$ can be computed as:
\begin{equation}
    \mu({\mathbf{x}}) = {\mathbf{k}}^\intercal(\mathbf{K} + \sigma_{\epsilon}^2 \mathbf{I}_N)^{-1} \mathbf{y},
    \label{eq:mean}
\end{equation}
\begin{equation}
    \sigma({\mathbf{x}}) = {\kappa}(\textbf{x},\textbf{x}) - {\mathbf{k}}^\intercal(\mathbf{K}+\sigma_{\epsilon}^2 \mathbf{I}_{N})^{-1}{\mathbf{k}},
    \label{eq:covariance}
\end{equation}
where ${\mathbf{k}}$ is the kernel vector between ${\mathbf{x}}$ and the data in $\mathcal{D}$, $\mathbf{K}$ is an $N \times N$ matrix with elements $\mathbf{K}_{p,q}=\kappa(\mathbf{x}^{(p)}, \mathbf{x}^{(q)})$, $p,q \in \{1,\ldots,N\}$, $\mathbf{I}_{N}$ is a $N \times N$ identity matrix, and $\mathbf{y}$ is the vector of noisy observations of the output function of interest.

A GP-based optimization pipeline leverages the uncertainty quantification capabilities of GPs to balance the exploitation and exploration of search spaces under unknown and expensive objective functions~\cite{andrea}, making it widely applicable across a plethora of real-world optimization problems~\cite{human-loop, gmm2}.
A pseudo-code for GP-based optimization is provided in \textbf{Algorithm \ref{alg: GP-UCB}}.
This optimization process proceeds iteratively, with each iteration generating a query solution using an acquisition function.
The procedure for obtaining the query solution in the $t$-th iteration can be expressed as:
\begin{equation}
\label{eq: acquistion}
    \mathbf{x}^{(t)} = \argmax_{\mathbf{x}\in \mathcal{X}} \alpha(\mathbf{x};\{(\mathbf{x}_i, f(\mathbf{x}_i))\}_{i=1}^{t + N_{init}}),
\end{equation}
where $\alpha(\sbullet)$ denotes the acquisition function. 
Many acquisition functions have been studied over the years, including expected improvement~\cite{aq-ei}, upper confidence bound~(UCB)~\cite{andrea}, knowledge gradient~\cite{aq-kg}, or entropy search~\cite{aq-es}.
\section{Parametric Multi-task Optimization with a Fixed Task Set}
In this section, we focus on a restricted version of the PMTO problem, considering only a finite and fixed set of optimization tasks $\{\mathcal{T}_m\}_{m=1}^{M}$ parameterized by task parameters $\{\boldsymbol{\theta}_m\}_{m=1}^{M}$. 
This configuration allows us to examine how the inclusion of task parameters in the optimization loop provides empirical and theoretical benefits over the single-task counterpart.
Since we consider solving optimization problems under limited evaluation budgets, the GP-based optimization in \textbf{Algorithm \ref{alg: GP-UCB}} serves as an established baseline method.
All subsequent analysis and discussions are therefore grounded in the framework of GP-based optimization.
It can be shown that by employing task parameters as side information in the GP models, we can inherently enable multi-task optimization and enhance the overall convergence performance for each task theoretically.
\subsection{Optimization of Fixed Parameterized Tasks}
As an instantiation of \textbf{Algorithm \ref{alg: GP-UCB}}, we employ the well-studied UCB\footnote{The formulation is given to solve a maximization problem but can also be adapted to the minimization problem by negation.}~\cite{andrea} acquisition function, defined as follows:
\begin{equation}
    \mathbf{x}^{(t)} = \argmax_{\mathbf{x}\in \mathcal{X}} -\mu(\mathbf{x}) + \beta \cdot \sigma(\mathbf{x}),
\label{eq: ucb}
\end{equation}
where the query solution $\mathbf{x}^{(t)}$ strikes a balance between exploitation and exploration in accordance with the trade-off coefficient $\beta$. 
The methodology is further extended to the PMTO framework with fixed tasks (termed as PMTO-FT), as illustrated in \textbf{Algorithm \ref{alg: UGP-UCB}}. 
Unlike common GP-based optimization, unified GP models are built in PMTO-FT using an augmented dataset $\mathcal{D}_{pmt} = \cup_{m=1}^{M} \{(\mathbf{x}_{m}^{(i)}, \boldsymbol{\theta}_m, f(\mathbf{x}_{m}^{(i)}, \boldsymbol{\theta}_m))\}_{i=1}^{N_{m}}$, where $N_m$ represents the number of samples evaluated for the $m$-th problem. 
This dataset encompasses not only the evaluated solutions and their corresponding objective function values but also the task parameters $\{\boldsymbol{\theta}_m\}_{m=1}^M$. 
The unified GP is thus trained to infer how a point in the product space $\mathcal{X} \times \Theta$ maps to the objective space.
Accordingly, the UCB acquisition function in PMTO-FT is defined as:
\begin{equation}
    \mathbf{x}_{m}^{(t)} = \argmax_{\mathbf{x}\in \mathcal{X}} -\mu_{pmt}(\mathbf{x}, \boldsymbol{\theta}_m) + \beta \cdot \sigma_{pmt}(\mathbf{x}, \boldsymbol{\theta}_m).
\label{eq: uni-ucb}
\end{equation}
Here $\mu_{pmt}(\mathbf{x}, \boldsymbol{\theta}_m)$ and $\sigma_{pmt}(\mathbf{x}, \boldsymbol{\theta}_m)$ are calculated as follows:
\begin{equation}
    \mu_{pmt}(\mathbf{x},\boldsymbol{\theta}_m) = {\mathbf{k}}_{pmt}^\intercal(\mathbf{K}_{pmt} + \sigma_{\epsilon}^2 \mathbf{I}_{N_{pmt}}^{-1}) \mathbf{y}_{pmt},
    \label{eq:mean_pmto}
\end{equation}
\begin{equation}
    \sigma_{pmt}(\mathbf{x},\boldsymbol{\theta}_m) = {\kappa}_{pmt,*} - \Tilde{\mathbf{k}}_{pmt}^\intercal(\mathbf{K}_{pmt}+\sigma_{\epsilon}^2 \mathbf{I}_{N_{pmt}})^{-1}{\mathbf{k}}_{pmt},
    \label{eq:covariance_pmto}
\end{equation}
where ${\kappa}_{pmt,*} = {\kappa}_{pmt}(({\mathbf{x}},\boldsymbol{\theta}_m),({\mathbf{x}},\boldsymbol{\theta}_m))$, ${\kappa}_{pmt}(\sbullet,\sbullet)$ is the unified kernel function, ${\mathbf{k}}_{pmt}$ is the kernel vector between $(\mathbf{x},\boldsymbol{\theta}_m)$ and the data in $\mathcal{D}_{pmt}$, $\mathbf{K}_{pmt}$ is the kernel matrix of the data in $\mathcal{D}_{pmt}$, $\mathbf{I}_{N_{pmt}}$ is an $N_{pmt} \times N_{pmt}$ identity matrix, $N_{pmt}$ is the size of $\mathcal{D}_{pmt}$, and $\mathbf{y}_{pmt}$ is the vector of noisy observations of the objective function values in $\mathcal{D}_{pmt}$.

\begin{algorithm}[!t]
\label{alg: UGP-UCB}
\DontPrintSemicolon
    \caption{PMTO-FT}
    \KwData{Initialization budget $N_{init}$, Total evaluation budgets $N_{tot}$, Objective function $f(\textbf{x},\bm{\theta})$, Target optimization tasks $\{\mathcal{T}_m\}_{m=1}^M$, and corresponding task parameters $\{\boldsymbol{\theta}_m\}_{m=1}^M$;} %
    \KwResult{The best solution found in solution set $\mathcal{D}_{pmt}$ for each optimization task;}
    $\mathcal{D}_{pmt} \gets \emptyset$ \;
    \ForEach{task $\mathcal{T}_m$}{
    Randomly initialize $N_{init}/M$ solutions\;
    Evaluate the initial solutions via the parameterized objective function $f(\sbullet, \boldsymbol{\theta}_m)$ of task $\mathcal{T}_m$\;
    $\mathcal{D}_{pmt} \gets \mathcal{D}_{pmt} \cup \{(\mathbf{x}_{m}^{(i)}, \boldsymbol{\theta}_m, f(\mathbf{x}_{m}^{(i)}, \boldsymbol{\theta}_m))\}_{i=1}^{N_{init}/M}$ \;
    }
    $t \gets 0$\;
    Train a unified GP model on $\mathcal{D}_{pmt}$\;
    \While{$t + N_{init}<N_{tot}$}
    {
    \ForEach{task $\mathcal{T}_m$}{
     Sample new input data $\mathbf{x}_{m, q}^{(t)}$ for the current optimization task based on (\ref{eq: uni-ucb})\;
     $\mathcal{D}_{pmt} \gets \mathcal{D}_{pmt} \cup \{(\mathbf{x}_{m}^{(t)}, \boldsymbol{\theta}_m, f(\mathbf{x}_{m}^{(t)}, \boldsymbol{\theta}_m))\}$ \;
    $t \gets t + 1$\;
    }
    Update the GP model according to $\mathcal{D}_{pmt}$\;
    }
\end{algorithm}

\subsection{Theoretical Analysis}\label{sec: theorem}
Extending the analysis technique in~\cite{andrea}, we define the \textit{instantaneous regret} of the $m$-th optimization task at the $t$-th evaluation as:
\begin{equation}
   r_m(t) = f_{m}(\mathbf{x}_{m}^{(t)}) - f_m(\mathbf{x}_{m}^{*}) \geq 0, \forall m \in [M],
\label{eq:instant}
\end{equation}
where $f_{m}(\sbullet) = f(\sbullet, \boldsymbol{\theta}_m)$ and $\mathbf{x}_m^* = \argmin_{\mathbf{x}\in \mathcal{X}} f_m(\mathbf{x})$.
Thereafter, the \textit{cumulative regret} of the $m$-th task over $T$ evaluations can be defined as:
\begin{equation}
  R_m(T) = \sum_{t=1}^T r_m(t), \forall m \in [M].
\label{eq:cumulative}
\end{equation}

Based on these definitions, a statistical bound on the cumulative regret for GP-based optimization procedures can be derived. 
This result holds under assumptions of a correctly specified GP prior, known additive noise variance $\sigma_{\epsilon}^2$, global optimization of the UCB acquisition function to ascertain $\textbf{x}_{m}^{(t)}$, and a finite decision space $\mathcal{X}$. 
According to \cite{andrea}, we have, 
\begin{equation}
    \mathbf{Pr}\{R_m(T) \leq \sqrt{C_1T\beta_T\gamma_{T,m}}~\forall T \geq 1 \} \geq 1 - \delta,
    \label{eq:bound}
\end{equation}
where $C_1 = 8/\log(1 + \sigma_{\epsilon}^{-2})$, $\beta_{T}$ is a predefined UCB coefficient dependent on the confidence parameter $\delta \in [0,1]$; $\gamma_{T,m}$ is the maximal information gain~(MIG) quantifying the maximal uncertainty reduction about the objective function $f_m$ from observing $T$ samples, which can be denoted as follows:
\begin{equation}
    \gamma_{T,m} = \max_{\textbf{x}_{m}^{(1)},\ldots,\textbf{x}_{m}^{(T)}} I([y_m^{(1)},\ldots,y_m^{(T)}];\textit{\textbf{f}}_{m}|\mathcal{D}_{pmt}).
\label{eq:gamma}
\end{equation}
Here $\textbf{x}_{m}^{(1)},\ldots,\textbf{x}_{m}^{(T)}$ are the samples corresponding to the $m$-th task, $\textit{\textbf{f}}_{m} = [ f_m(\textbf{x}_m^{(1)}), \ldots, f_m(\textbf{x}_m^{(T)}) ]$,  $[y_{m}^{(1)},\ldots,y_{m}^{(T)}]$ are the observed noisy outputs, and $I$ represents the mutual information between $\textit{\textbf{f}}_{m}$ and $[y_{m}^{(1)},\ldots,y_{m}^{(T)}]$. 
From \eqref{eq:bound}, we see that the cumulative regret bound is determined primarily by the MIG. Thus, comparing the MIG under PMTO-FT with that for separately and independently solved tasks shall translate to a theoretical comparison of their respective convergence rates. A lower MIG implies tighter regret bounds and hence suggests faster convergence.
Denoting the MIG of the independent strategy as $\gamma_{T,m}^{ind}$, we establish the following theorem.

\begin{theorem}\footnote{A proof is provided in the supplementary material. Related theoretical results are also available in the literature~\cite{nips-ong, dai2020multitask}.}
    Let the kernel functions used in PMTO-FT and the independent strategy satisfy $\kappa_{pmt}((\mathbf{x}, \boldsymbol{\theta}),(\mathbf{x}', \boldsymbol{\theta}))=\kappa_{ind}(\mathbf{x},\mathbf{x}')$. Then, the MIG in the independent strategy, denoted as $\gamma_{T,m}^{ind}$, and the MIG in the PMTO-FT, denoted as $\gamma_{T,m}$, satisfy $\gamma_{T,m} \leq \gamma_{T,m}^{ind}, \forall m \in [M], \forall T \geq 1$.
\end{theorem}

\begin{remark}
    Given an evaluation budget of $(M \times T)$, each objective function $f_m(\sbullet), \forall m \in [M]$, is evaluated $T$ times in both PMTO-FT and the independent GP-based optimization procedure. This ensures a fair distribution of the computational budget across tasks while comparing the cumulative regret bounds of the two algorithms.
\end{remark}

A detailed proof is in the appendix.
According to the result, incorporating task parameters $\{\boldsymbol{\theta}_m\}_{m=1}^{M}$ as side information results in a lower MIG for the PMTO-FT method compared to the independent strategy.
This translates to faster convergence in PMTO-FT relative to its single-task counterpart.
Furthermore, due to the explicit modeling of inter-task relationships through the unified GP kernel, PMTO-FT's approximation model can potentially generalize to tasks beyond the fixed training set of $M$ optimization tasks.
This attribute positions PMTO-FT as a promising approach for extension to address the problem defined by formulations~(\ref{eq: parameterized-mto}) and (\ref{eq: task_model}).

\section{The ($\boldsymbol{\theta}_l, \boldsymbol{\theta}_u$)-PMTO Algorithm}

In this section, we present the $(\boldsymbol{\theta}_l, \boldsymbol{\theta}_u)$-PMTO algorithm that relaxes PMTO-FT's constraint of a fixed task set. The proposed method integrates a probabilistic task model that fulfills two key roles.
First, it learns to predict optimized solutions for parameterized tasks within the continuous and bounded task space, thereby broadening optimization capacity from a fixed set to the entire task space.
Second, the inter-task relationships captured by the model help guide an evolutionary search over under-explored regions of the task space---a process termed \textit{task evolution}---supporting the acquisition of informative data that enhances the performance of the approximation models. 

\subsection{Overview of the $(\boldsymbol{\theta}_l, \boldsymbol{\theta}_u)$-PMTO}
A pseudo-code of the proposed $(\boldsymbol{\theta}_l, \boldsymbol{\theta}_u)$-PMTO is detailed in \textbf{Algorithm \ref{alg: GP-EASS}} and its key steps are explained below.
\begin{itemize}
    \item \textit{Initialization}: $M$ parameterized optimization tasks are initiated by randomly sampling a set of task parameters $\Psi = \{ \boldsymbol{\theta}_{m} \}_{m=1}^{M}$ in the task space. $N_{init}/M$ solutions are initialized and evaluated for each task, to form dataset $\mathcal{D}_{pmt}$. A dataset $\mathcal{D}^*$, which includes only the current best evaluated samples corresponding to all task parameters in $\Psi$, is also induced. 
    \item \textit{Approximation Models}: In steps \textbf{9} to \textbf{10} and steps \textbf{21} to \textbf{23}, a unified GP model (with inputs comprising both the solutions and the task parameters) and a task model $\mathcal{M}$ are built based on dataset $\mathcal{D}_{pmt}$ and $\mathcal{D}^*$, respectively.
    The unified GP approximates the mapping from the product space $\mathcal{X} \times \Theta$ to the objective space, while the task model maps points in the task space $\Theta$ to their corresponding elite solutions.
    \item \textit{Task Evolution}: In steps \textbf{13} to \textbf{14}, the inter-task relationships captured by the task model $\mathcal{M}$ are used to guide an EA to search for new tasks to be included in the task pool $\Psi$. The details of the task evolution module can be found in Section V-C.
    \item \textit{Cross-task Optimization}: In lines \textbf{16} to \textbf{20}, the predictive distribution of the unified GP is used to define acquisition functions, formulated in (\ref{eq: uni-ucb}), whose optimization yield potentially fit solution candidates for each task in $\Psi$. 
\end{itemize}
At the end of a full ($\boldsymbol{\theta}_l$, $\boldsymbol{\theta}_u$)-PMTO run, the algorithm returns the best solutions found for each task in the pool $\Psi$, and the resultant task model $\mathcal{M}$.
Note that in lines \textbf{16} to \textbf{20}, the optimization of all tasks in the pool progresses in tandem. This multi-task approach lends an advantage in terms of efficient coverage of the task space as compared to a sequential strategy where sampled tasks are optimized one after another~\cite{RAL-CBO}.

The effectiveness of PMTO is strongly contingent upon the quality of the task model, which depends not only on the convergence of the offline optimization to high-quality solutions, but also on the coverage and placement of sampled tasks in the task space.
Simply optimizing over the initial task set or randomly sampled tasks may fail to capture regions of the task space characterized by high variation in the task-to-solution mapping~\cite{inverse-density}.
To address this, we introduce a task evolution module detailed in Section V-C, which strategically selects new tasks during the offline optimization step of the ($\boldsymbol{\theta}_l$, $\boldsymbol{\theta}_u$)-PMTO algorithm.
The impact of this design is further
verified through experiments in Section VI-C.

\begin{algorithm}[!t]
\label{alg: GP-EASS}
\DontPrintSemicolon
    \caption{$(\boldsymbol{\theta}_l, \boldsymbol{\theta}_u)$-PMTO}
    \KwData{Initial task size $M$, Initialization budget $N_{init}$, Total budgets for each task $N_{tot}$, Objective function $f(\textbf{x},\bm{\theta})$, Evaluated solution set $\mathcal{D}_{pmt}$, Task pool $\Psi$;}
    \KwResult{Best solution found for each optimization task;}
    $\mathcal{D}_{pmt} \gets \emptyset$ \;
    $\Psi \gets \{\boldsymbol{\theta}_m\}_{m=1}^{M}$\;
    \ForEach{task parameter in $\Psi$}{
    Randomly initialize $N_{init}/M$ solutions\;
    Evaluate the objective function for the initial solutions via $f(\sbullet, \boldsymbol{\theta}_m)$\;
    $\mathcal{D}_{pmt} \gets \mathcal{D}_{pmt} \cup \{(\mathbf{x}_{m}^{(i)}, \boldsymbol{\theta}_m, f(\mathbf{x}_{m}^{(i)}, \boldsymbol{\theta}_m))\}_{i=1}^{N_{init}/M}$ \;
    }
    Let $\mathbf{x}_{m}^{*}$ be the current best solution corresponding to the optimization task parameterized by $\boldsymbol{\theta}_m$, then $\mathcal{D}^* \gets \{(\boldsymbol{\theta}_m,  \mathbf{x}_{m}^{*})\}_{m=1}^{M}$\;
    Train a unified GP model on $\mathcal{D}_{pmt}$\;
    Train the task model $\mathcal{M}$ on $\mathcal{D}^*$\;
    $t \gets 0$\;
    \While{$t + N_{init}<N_{tot}$}
    {
    $\boldsymbol{\theta}_{new} \gets$ Task-Evolution($\mathcal{M}$, $\Psi$, $P$, $G$)\;
    $\Psi \gets \Psi \cup \{\boldsymbol{\theta}_{new}\}$\;
    $M \gets M + 1$ \;
    \ForEach{task parameter $\boldsymbol{\theta}_m$ in $\Psi$}{
    Sample new input data $\mathbf{x}_{m}^{(t)}$ for the current optimization task based on (\ref{eq: uni-ucb})\;
    $\mathcal{D}_{pmt} \gets \mathcal{D}_{pmt} \cup \{(\mathbf{x}_{m}^{(t)}, \boldsymbol{\theta}_m, f_{m}(\mathbf{x}_{m}^{(t)}, \boldsymbol{\theta}_m))\}$ \;
    $t \gets t + 1$
    }
    Update the unified GP model on $\mathcal{D}_{pmt}$\;
    Let $\mathbf{x}_{m}^{*}$ be the current best solution corresponding to the $m$-th task parameters in $\Psi$, then $\mathcal{D}^* \gets \{(\boldsymbol{\theta}_m, \mathbf{x}_{m}^{*})\}_{m=1}^{M}$\;
    Update the task model $\mathcal{M}$ on $\mathcal{D}^*$\;
    }
\end{algorithm}

\subsection{GP-based Task Model}
The dataset $\mathcal{D}^* = \{(\boldsymbol{\theta}_m,\mathbf{x}_{m}^{*})\}_{m=1}^{M}$, where $\mathbf{x}_{m}^{*}$ indicates the best solution found so far for parametrized task $\boldsymbol{\theta}_m$, can be used to approximate a mapping from tasks to their optimized solutions.
We instantiate this task model by concatenating multiple independent GPs.
Specifically, assuming $\mathcal{X} \subset \mathbb{R}^V$, $V$ GPs are built with each model corresponding to a separate dimension of the solution space.
For the $v$-th dimension, a data subset is constructed as $\mathcal{D}_v^* \coloneqq \{(\boldsymbol{\theta}_m, \mathbf{x}_{m,v}^{*})\}_{m=1}^{M}$.
Given $\mathcal{D}_v^*$, the posterior estimate of the $v$-th decision variable value, $\mathcal{M}_{v}(\boldsymbol{\theta}) = \mathcal{N}(\Tilde{\mu}_v(\boldsymbol{\theta}), \Tilde{\sigma}_v(\boldsymbol{\theta}))$, for a queried optimization task $\boldsymbol{\theta}$ is given as:
\begin{equation}
    \Tilde{\mu}_v(\boldsymbol{\theta}) = \Tilde{\mathbf{k}}_v^\intercal(\Tilde{\mathbf{K}}_v + \sigma_{\epsilon, v}^2 \mathbf{I}_M)^{-1} \mathbf{x}_v,
    \label{eq:task_mean}
\end{equation}
\begin{equation}
    \Tilde{\sigma}_v(\boldsymbol{\theta}) = \Tilde{\kappa}_v(\boldsymbol{\theta}, 
    \boldsymbol{\theta}) - \Tilde{\mathbf{k}}_v^\intercal(\Tilde{\mathbf{K}}_v+\sigma_{\epsilon, v}^2 \mathbf{I}_{M})^{-1}\Tilde{\mathbf{k}}_v,
    \label{eq:task_covariance}
\end{equation}
where $\Tilde{\kappa}_v(\sbullet, \sbullet)$ is the kernel function used in the $v$-th model, $\Tilde{\mathbf{k}}_v$ denotes the kernel vector between parameters of the queried task and the existing task parameters in $\Psi$, $\Tilde{\mathbf{K}}_v$ is the overall kernel matrix of the $v$-th GP, $\mathbf{x}_v$ is a column vector containing the $v$-th variable of all solutions in dataset $\mathcal{D}_v^*$, and $\sigma_{\epsilon, v}$ is the noise term of the $v$-th GP.
The concatenated output of the task model is then given as $\mathcal{M}(\boldsymbol{\theta}) = [ \mathcal{M}_1(\boldsymbol{\theta}), \ldots, \mathcal{M}_{V}(\boldsymbol{\theta}) ]^T$, such that $\mathcal{M}(\boldsymbol{\theta})$ lies in the solution space $\mathcal{X} \subset \mathbb{R}^V$.

\begin{algorithm}[!t]
\label{alg: Task-Evolution}
\DontPrintSemicolon
    \caption{Task Evolution}
    \KwData{Task pool $\Psi$, Task model $\mathcal{M}$, Population size $P$, Maximum generations $G$;}
    \KwResult{New task parameter $\boldsymbol{\theta}_{new}$;}
    Initialize a population $\{{\boldsymbol{\theta}}_0^{(p)}\}_{p=1}^P$ of candidate task parameters\;
    Evaluate scores $g({\boldsymbol{\theta}}_0^{(p)})$ using equation (\ref{eq: objective-function})\;
    \For{$\tau = 1$ \KwTo $G$}{
        Select parents using binary tournament selection\;
        Generate offspring with SBX crossover\;
        Apply PM to mutate offspring\;
        Evaluate $g({\boldsymbol{\theta}})$ for offspring individuals;
        Combine parent and offspring populations\;
        Select top $P$ individuals as $\{{\boldsymbol{\theta}}_{\tau}^{(p)}\}_{p=1}^P$ based on $g({\boldsymbol{\theta}})$ scores;
    }
    Return ${\boldsymbol{\theta}}_{new} = \argmax_{\boldsymbol{\theta} \in \{{\boldsymbol{\theta}}_{G}^{(p)}\}_{p=1}^P} g({\boldsymbol{\theta}})$\;
\end{algorithm}
\subsection{Task Evolution}
The task evolution module plays a central role in improving task space coverage with the ultimate goal of enhancing the performance of PMTO in online operation. 
Instead of relying on simple strategies such as random sampling or optimizing only on the initial task set, task evolution is guided by the iteratively updated task model to actively select the placement of tasks during offline optimization. This guided selection steers exploration toward under-explored and complex regions of the task space (such as those with high variation in the task-to-solution mapping), thereby enhancing the task model's ability to predict optimized solutions.
If tasks are sampled without careful consideration, poor coverage of the task space may occur. 
A direct comparison of task evolution versus the random sampling of tasks is presented in the numerical studies in Section VI.

To build a well-calibrated task model, it is essential to sample diverse tasks, ensuring a good coverage of the task space. 
This is accomplished by searching for $\boldsymbol{\theta}$ that maximizes the following objective function:
\begin{equation}
    g(\boldsymbol{\theta}) = \Sigma_{v=1}^V\det Q_v([\boldsymbol{\theta}_1, \boldsymbol{\theta}_2, \ldots, \boldsymbol{\theta}_M, \boldsymbol{\theta}]),
\label{eq: objective-function}
\end{equation}
where $Q_v$ is a $(M+1)\times(M+1)$ matrix with element $Q_{v_{i,j}} = \Tilde{\kappa}_v(\boldsymbol{\theta}_i, \boldsymbol{\theta}_j)$, and $\boldsymbol{\theta}_i, \boldsymbol{\theta}_j \in \{\boldsymbol{\theta}_m\}_{m=1}^M \cup \{ \boldsymbol{\theta} \}$. 
The $(i,j)$-th element of the matrix thus captures the inter-task relationship between task pair $\boldsymbol{\theta}_i$ and $\boldsymbol{\theta}_j$ as defined by the $v$-th kernel function.
This formulation is inspired by Determinantal Point Processes (DPP)~\cite{dpp, effective-diversity}, where the determinant of a kernel matrix quantifies the overall diversity of a set of items. Intuitively, a larger determinant value reflects that the selected tasks are more widely spread in the kernel-induced space, indicating lower similarity and higher diversity among them~\cite{dpp}. 
In this paper, this property promotes the selection of tasks that are less well represented by the current task model.
This is particularly crucial in PMTO, where limited evaluation budgets may otherwise lead to redundant task sampling and poor coverage of the task space.
By maximizing $g(\boldsymbol{\theta})$, the task evolution module actively guides the task selection towards under-explored and complex regions of the task model, where the mapping from task parameters to solutions may be less certain.
Purely maximizing this determinant-based coverage objective over a set of task parameters can be NP-hard~\cite{dpp-properties}. However, taking the logarithm of each determinant yields a monotone, submodular set function~\cite{dpp-properties}, which admits a greedy, iterative maximization procedure that achieves a $(1-1/e)$-approximation to the optimal coverage~\cite{submodular}. 
This property motivates the design of the task evolution module, which aims to iteratively improve task-space coverage. 
Since deriving gradients on $g(\boldsymbol{\theta})$ would require expensive and numerically fragile backpropagation through multiple kernel-matrix inversions and determinant computations, the task evolution module instead employs derivative-free evolutionary algorithms to optimize the objective function in (\ref{eq: objective-function}).

To optimize the objective function $g(\boldsymbol{\theta})$ in each iteration of $(\boldsymbol{\theta}_{l},\boldsymbol{\theta}_{u})$-PMTO, we adopt the simple EA described in \textbf{Algorithm \ref{alg: Task-Evolution}}.
The EA incorporates polynomial mutation~(PM)~\cite{pm} and simulated binary crossover~(SBX)~\cite{sbx}.
Binary tournament selection is employed to impart selection pressure for evolving tasks from one generation to the next.
Note that while our algorithmic choice is motivated by simplicity, other evolutionary methodologies could be readily used for searching the optimal placement of tasks.

\begin{figure*}[!t]
\centering
\includegraphics[width=5.7in]{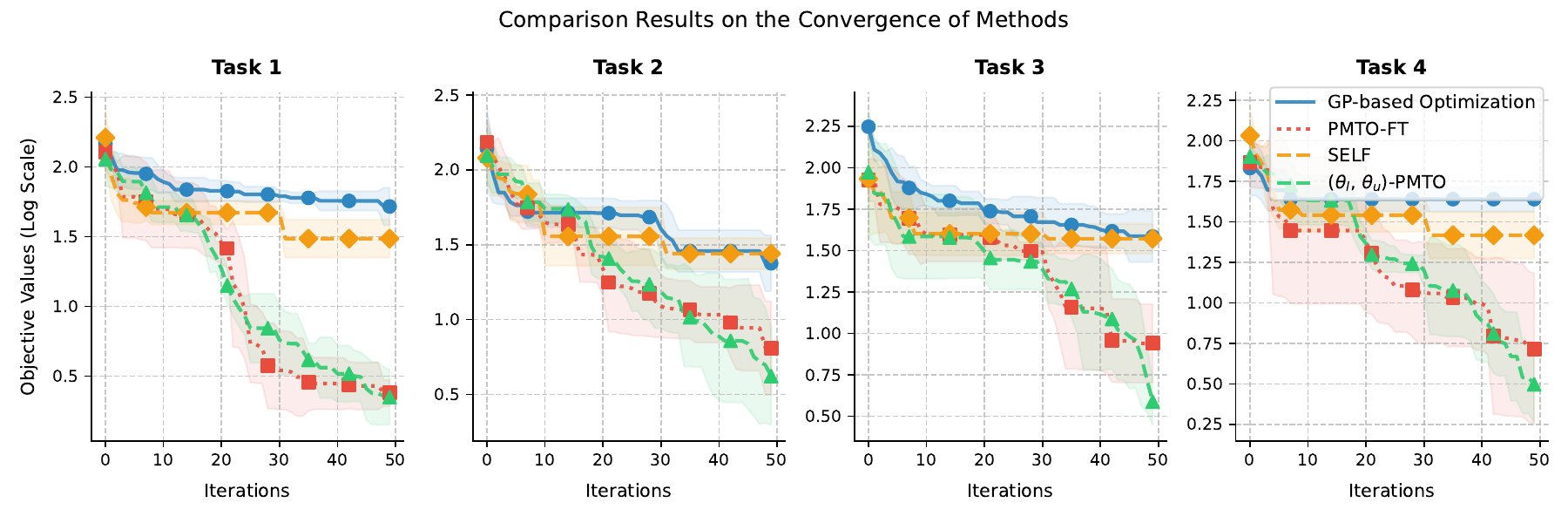} \\[1em] 
\includegraphics[width=5.7in]{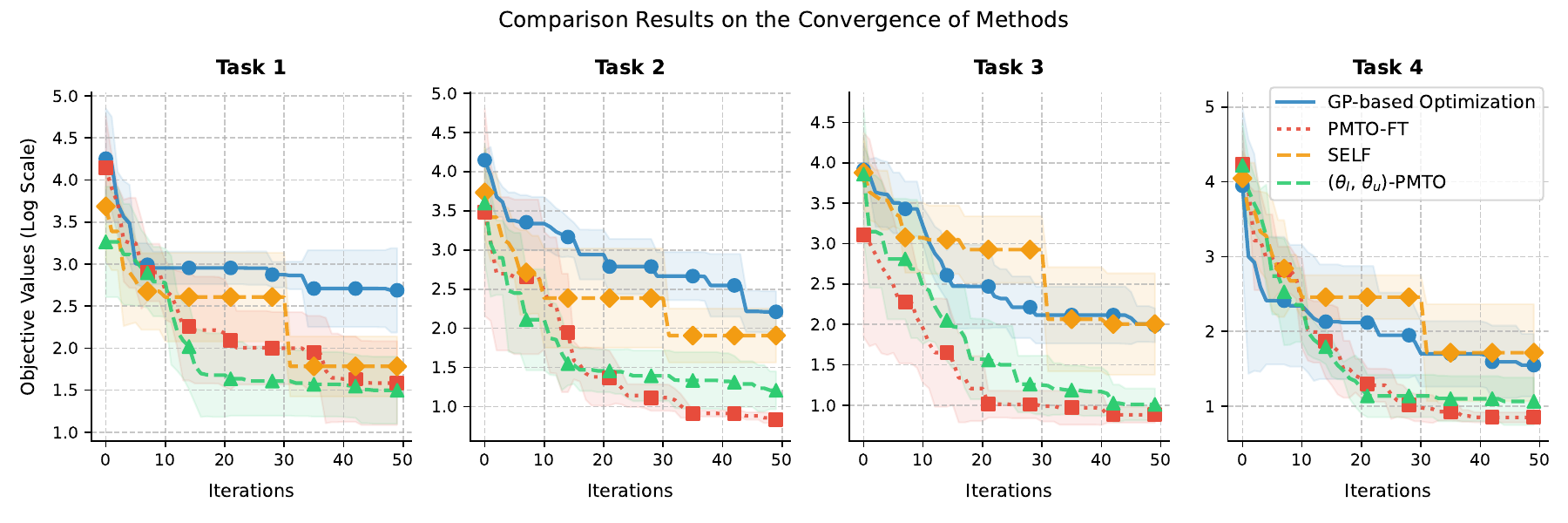}
\caption{Convergence trends in the offline optimization stage for Ackley-II (top) and Griewank-II (bottom) on four sample tasks each.}
\label{Fig: combined-convergence}
\end{figure*}

\section{Results}
\subsection{Experimental Settings}
In this section, we assess the effectiveness of the proposed $(\boldsymbol{\theta}_l, \boldsymbol{\theta}_u)$-PMTO on various PMTO problems, encompassing synthetic problems, three adaptive control problems, and a robust engineering design problem.
We compare $(\boldsymbol{\theta}_l, \boldsymbol{\theta}_u)$-PMTO to GP-based optimization introduced in \textbf{Algorithm \ref{alg: GP-UCB}}, PMTO-FT introduced in \textbf{Algorithm \ref{alg: UGP-UCB}} and a recent expensive MTO algorithm, SELF~\cite{SELF}.
All algorithms except SELF~(employing expected improvement in its implementation) employ the UCB as the acquisition function, with $\beta$ set to 1.0 in (\ref{eq: ucb}) and (\ref{eq: uni-ucb}).
Across all problems, the total evaluation budget $N_{tot}$ is 2000 and the initialization budget $N_{init}$ is 200.
For GP-based optimization and PMTO-FT, the target optimization tasks are randomly sampled in the task space via Latin hypercube sampling~\cite{lhs} with the sample size $M$ of 20.
The GP models applied in all the methods are configured with the same hyperparameters.
In this paper, we instantiate the GP model via RBF kernels~\cite{GPML}.
To optimize the hyperparameters of the GP, the Adam optimizer~\cite{adam} with a learning rate of 0.01 and a maximum epoch of 500 is employed.
For the task evolution module, P and G are set to 100 and 50, respectively, the SBX crossover operator is parameterized by distribution index $\eta_c=15$ and $p_c=0.9$, and the PM mutation operator is parameterized by distribution index $\eta_m=20$ and $p_m=0.9$.

\begin{figure*}[!t]
\centering
\subfloat[]{\includegraphics[width=2.2in]{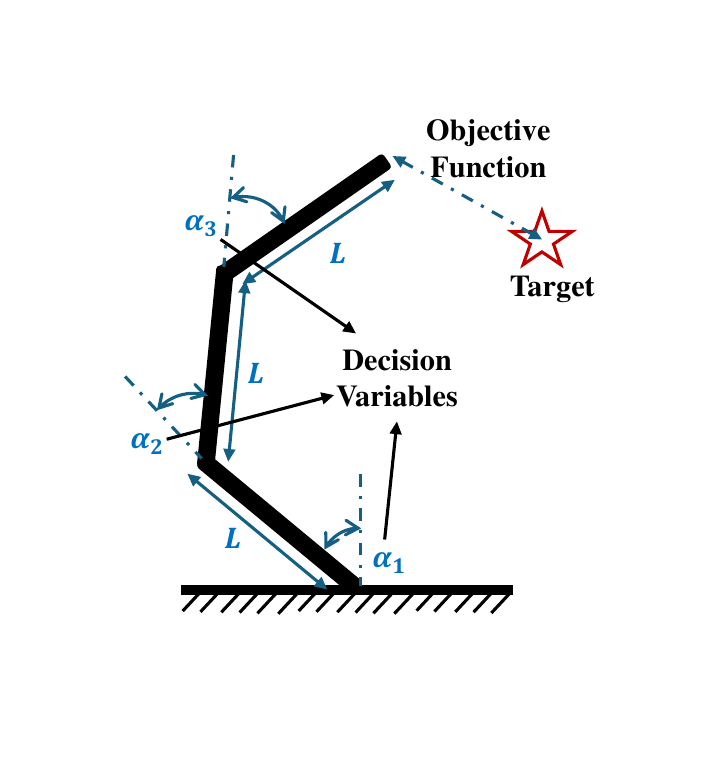}%
\label{fig: real1}}
\subfloat[]{\includegraphics[width=2.8in]{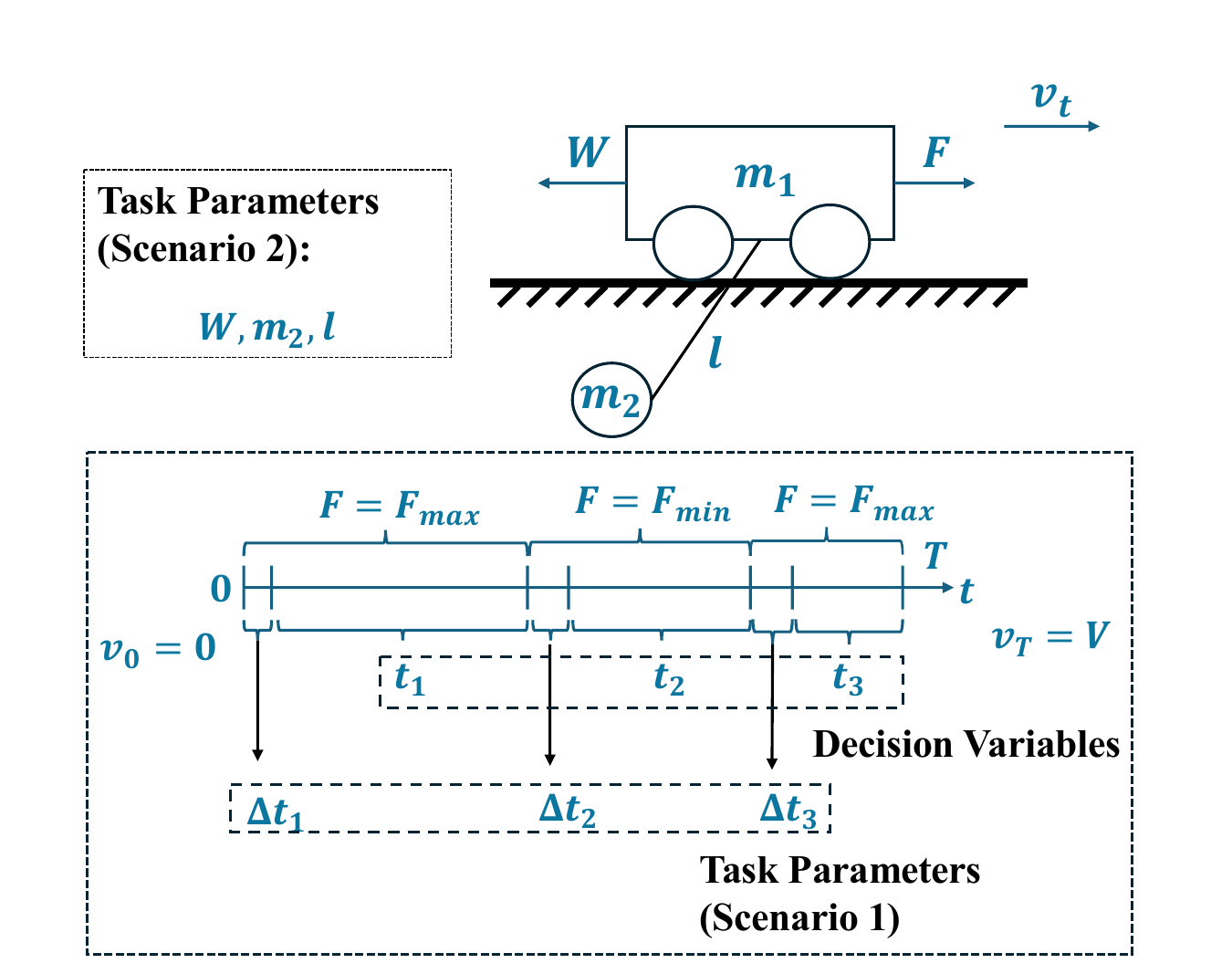}%
\label{fig: real2}}
\subfloat[]{\includegraphics[width=2.3in]{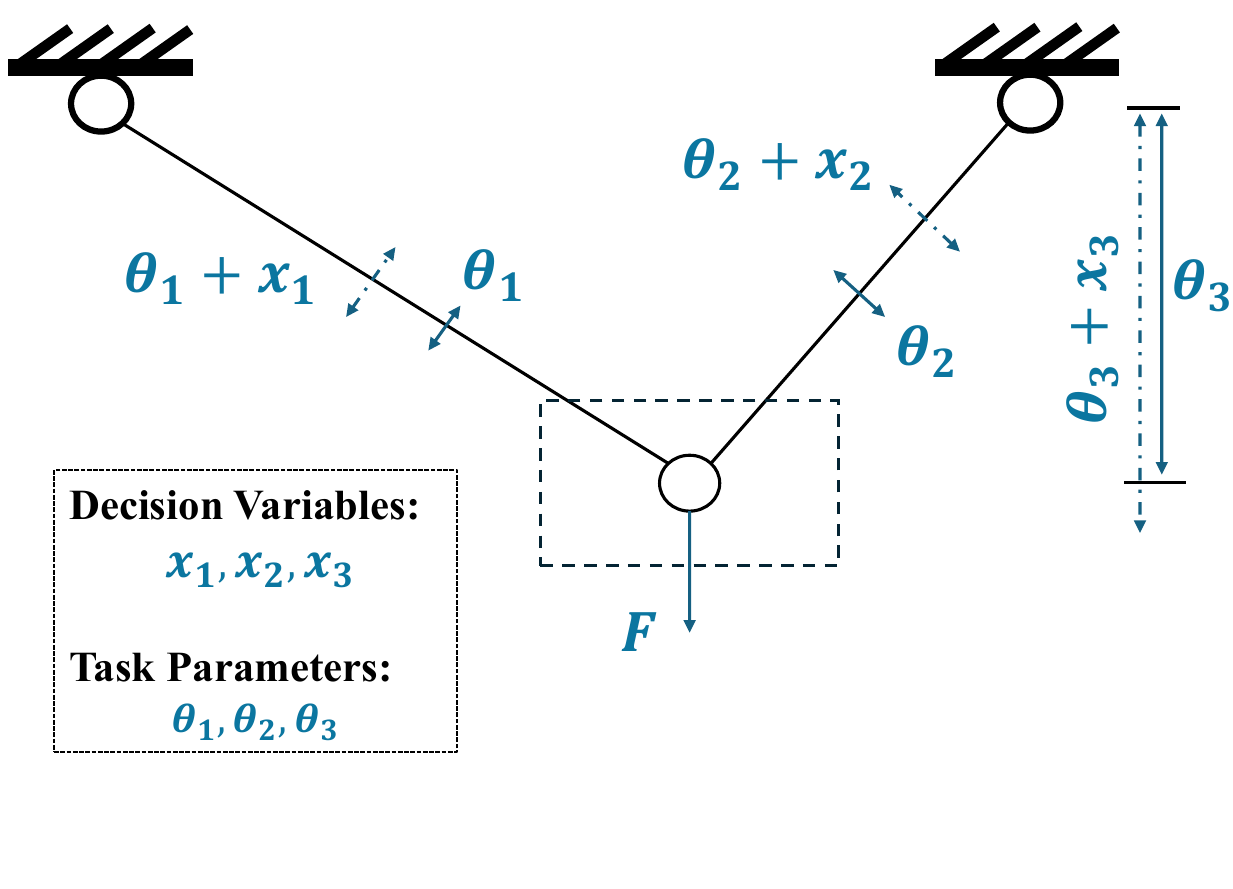}
\label{fig:real3}}%
\hfil
\caption{The illustrative examples for the case studies: (a) Parametric Robot Arm Optimization: Optimize the angular position of each joint ($\alpha_1, \alpha_2, \alpha_3$) so that the end effector can approach as close as to the target position. Task parameters include the length of each arm $L$ and the maximum rotation degree $\alpha_{max}$. (b) Parametric Crane-Load System Optimization: Optimize the time intervals $t_1, t_2, t_3$ during which the distinct drive forces $F$ are exerted on the crane-load system so that the system can achieve a goal velocity with minimal operating time and oscillation. Task parameters include Scenario 1: time delays $\Delta t_1, \Delta t_2, \Delta t_3$, and Scenario 2: operating conditions including the length of suspension $l$, the mass of load $m_2$, and the resistance $W$. (c) Plane Truss Design: Optimize the cross-sectional areas of two bars $\theta_1, \theta_2$ and $\theta_3$ (the vertical distance from the second bar) to minimize the overall structural weight and the joint displacement in aware of possible processing errors $x_1, x_2, x_3$. The exact formulation of the (b) and (c) can be found in the supplementary materials.}
\label{fig: real}
\end{figure*}

\subsection{Performance Metrics}
To assess the performance of the proposed method, we conduct $U = 20$ independent trials of experiments for each algorithm, and numbers with indicators ($+$), ($\mathbf{-}$) and ($\mathbf{\approx}$) imply that the compared algorithm is better than, worse than, or similar to the proposed $(\boldsymbol{\theta}_l, \boldsymbol{\theta}_u)$-PMTO at 90\% confidence level as per the Wilcoxon signed-rank test. 
Since GP-based optimization, PMTO-FT, and SELF lack task models capable of addressing infinite parameterized tasks, for fairness, task models are constructed offline for these algorithms as part of the evaluation process.
For GP-based optimization, $M$ optimization tasks are solved independently.
For PMTO-FT, the same $M$ optimization tasks are solved collaboratively with the unified GP model in the product space $\mathcal{X}\times\Theta$.
For SELF, multi-task GP models solve $M$ optimization tasks without taking into account the task parameterization.
After the optimization process, the best solutions obtained for each task constitute a dataset $\mathcal{D}^* = \{(\boldsymbol{\theta}_m,\mathbf{x}_{m}^{*})\}_{m=1}^{M}$ as in the proposed PMTO framework, and task models for GP-based optimization, PMTO-FT, and SELF are trained on the corresponding dataset as formulated in \eqref{eq:task_mean} and \eqref{eq:task_covariance}.
For evaluation, a set of randomly sampled task parameters is provided to the task model, which then generates a set of predicted solutions (simulating online operation). 
These solutions are subsequently evaluated via the objective function and ranked based on their objective values. 
Quantiles are recorded at the 5th, 25th, 50th, 75th, and 95th percentiles to assess solution quality across sampled tasks. 
The mean value for each quantile is then computed across all trials to evaluate the performance distribution.

To define the evaluation metrics, let the set of sampled task parameters be $\bar{\Theta} = \{\boldsymbol{\theta}_1, \boldsymbol{\theta}_2, \ldots, \boldsymbol{\theta}_K\}$.
For each trial $u \in [U]$, the task model $\mathcal{M}(\boldsymbol{\theta}_k)$ predicts the optimized solution corresponding to each task parameter $\boldsymbol{\theta}_k$\footnote{For the final obtained task model of $(\boldsymbol{\theta}_l, \boldsymbol{\theta}_u)$-PMTO, only top $p$\% of the near-optimal solutions are employed to train the task model, where $p$ is set to $70$ in this paper.}.
The associated optimization performance can be denoted as:
\begin{equation}
    F_u(\boldsymbol{\theta}_k) = f(\mathcal{M}(\boldsymbol{\theta}_k), \boldsymbol{\theta}_k), \forall u \in [U]
\end{equation}
Then, the optimization results in $\{ F_u(\boldsymbol{\theta}_k)\}_{k=1}^{K}$ are ranked, and the quantile values are computed as follows:
\begin{equation}
    P_{\alpha, u}=Quantile_{\alpha}(F_u(\boldsymbol{\theta}_k);\boldsymbol{\theta}_k\in \bar{\Theta})
\end{equation}
where $\alpha \in \{0.05, 0.25, 0.50, 0.75, 0.95\}$.
The final performance metric for each quantile is the mean value across all trials:
\begin{equation}
    \bar{P}_\alpha = \frac{1}{U}\sum_{u=1}^{U}P_{\alpha, u}, \alpha \in \{0.05, 0.25, 0.50, 0.75, 0.95\}.
\end{equation}
This metric provides a detailed understanding of the optimization performance of the task model across all the sampled task parameters. 
In this paper, $K$ is set to $100^2$ for task parameters in 2-dimensional space, $K$ is set to $10^5$ for task parameters in 5-dimensional space and $K$ is set to $V^{20}$ for the other $V$-dimensional task parameter spaces.

\subsection{Results on Synthetic Problems}

We assess the effectiveness of our methods using synthetic problems based on canonical objective functions modified with task parameters.
Specifically, the objective function is defined as:
\begin{equation}
    f(\mathbf{x}, \boldsymbol{\theta}) = g(\lambda(\mathbf{x}-\boldsymbol{\sigma}(L\boldsymbol{\theta})))
\end{equation}
where $g(\sbullet)$ is a base objective function, such as continuous optimization functions including Sphere, Ackley, Rastrigin, and Griewank to model the optimization problem, $\mathbf{x}$ represents the decision variable within a $V$-dimensional space, $\boldsymbol{\theta}$ denotes the task parameter within a $D$-dimensional space, $\lambda > 0$ is a scaling factor to adjust the magnitude of decision variables, $L \in \mathbb{R}^{V \times D}$ is a linear transformation matrix that maps task parameters into the $V$-dimensional space, and $\boldsymbol{\sigma}$ represents a nonlinear transformation applied to the transformed task parameter.
Totally, we construct eight synthetic problems: Sphere-I, Sphere-II, Ackley-I, Ackley-II, Rastrigin-I, Rastrigin-II, Griewank-I, and Griewank-II. 
Comprehensive details about these problems are provided in Section S-I of the supplementary materials.

\begin{table*}
\centering
\caption{Comparative Results of the Task Model's Online Performance. The results include Single-task Baseline Method, MTO Method, and PMTO variants (with or without task evolution) on Synthetic Test Problems. $U=20$ independent trials are considered.}
\resizebox{18cm}{!}  
{  
\begin{tabular}{c|c|c|c|c|c|c}
\hline
Problems                      & Quantile & GP-based Optimization   & SELF                              & PMTO-FT                          & $(\boldsymbol{\theta}_l, \boldsymbol{\theta}_u)$-PMTO-RT                         & $(\boldsymbol{\theta}_l, \boldsymbol{\theta}_u)$-PMTO                             \\ \hline\hline
\multirow{5}{*}{Sphere-I}     & 5\%      & 3.4605e-02 (2.2632e-02)~$\mathbf{\approx}$ & 3.5375e-02 (1.6824e-02)~$\mathbf{\approx}$ & \textbf{3.3402e-02 (2.6492e-02)}~$\mathbf{\approx}$ & 8.7646e-02 (6.5269e-02)~$\mathbf{-}$          & 4.3264e-02 (1.6935e-02)          \\ \cline{2-7} 
                              & 25\%     & 1.2633e-01 (7.6585e-02)~$\mathbf{\approx}$ & 1.5640e-01 (8.1991e-02)~$\mathbf{\approx}$ & \textbf{1.2448e-01 (1.1445e-01)}~$\mathbf{\approx}$ & 2.7083e-01 (2.2507e-01)~$\mathbf{-}$          & 1.2737e-01 (5.0277e-02)          \\ \cline{2-7} 
                              & 50\%     & 2.9818e-01 (1.5592e-01)~$\mathbf{\approx}$ & 4.3039e-01 (1.7511e-01)~$\mathbf{-}$ & 3.1267e-01 (2.8268e-01)~$\mathbf{\approx}$          & 5.8800e-01 (5.7331e-01)~$\mathbf{-}$          & \textbf{2.4507e-01 (1.0836e-01)} \\ \cline{2-7} 
                              & 75\%     & 8.9086e-01 (3.0498e-01)~$\mathbf{-}$ & 1.0883e+00 (3.1372e-01)~$\mathbf{-}$ & 6.5575e-01 (3.6466e-01)~$\mathbf{-}$          & 1.2854e+00 (1.2831e+00)~$\mathbf{-}$          & \textbf{4.3753e-01 (2.3145e-01)} \\ \cline{2-7} 
                              & 95\%     & 3.4692e+00 (6.0544e-01)~$\mathbf{-}$ & 3.6327e+00 (4.5250e-01)~$\mathbf{-}$ & 2.5749e+00 (1.0022e+00)~$\mathbf{-}$          & 3.3264e+00 (2.8819e+00)~$\mathbf{-}$          & \textbf{8.3653e-01 (4.1045e-01)} \\ \hline\hline
\multirow{5}{*}{Sphere-II}    & 5\%      & 1.2980e+00 (8.1092e-02)~$\mathbf{-}$  & 1.3832e+00 (2.1743e-02)~$\mathbf{-}$ & 1.3668e+00 (4.8590e-02)~$\mathbf{-}$           & 1.0247e+00 (2.8961e-01)~$\mathbf{-}$          & \textbf{4.2056e-01 (2.5906e-01)}  \\ \cline{2-7} 
                              & 25\%     & 2.2877e+00 (8.6464e-02)~$\mathbf{-}$  & 2.2184e+00 (2.2493e-03)~$\mathbf{-}$ & 2.1996e+00 (2.2345e-02)~$\mathbf{-}$           & 1.9613e+00 (2.8010e-01)~$\mathbf{-}$          & \textbf{1.1867e+00 (5.3394e-01)}  \\ \cline{2-7} 
                              & 50\%     & 3.2896e+00 (1.7396e-01)~$\mathbf{-}$  & 3.0538e+00 (2.8879e-02)~$\mathbf{-}$ & 3.0190e+00 (5.2262e-02)~$\mathbf{-}$           & 2.8556e+00 (2.7824e-01)~$\mathbf{-}$          & \textbf{2.0908e+00 (7.1504e-01)}  \\ \cline{2-7} 
                              & 75\%     & 4.5694e+00 (3.5324e-01)~$\mathbf{-}$  & 4.0529e+00 (5.4464e-02)~$\mathbf{\approx}$ & 4.0215e+00 (1.0754e-01)~$\mathbf{\approx}$           & 3.9965e+00 (3.1599e-01)~$\mathbf{\approx}$          & \textbf{3.3347e+00 (1.0237e+00)}  \\ \cline{2-7} 
                              & 95\%     & 6.9576e+00 (1.0843e+00)~$\mathbf{\approx}$  & \textbf{5.5961e+00 (5.3679e-02)}~$\mathbf{\approx}$ & 5.6836e+00 (2.4876e-01)~$\mathbf{\approx}$           & 6.0923e+00 (5.4639e-01)~$\mathbf{\approx}$          & 5.6729e+00 (2.1257e+00)  \\ \hline\hline
\multirow{5}{*}{Ackley-I}     & 5\%      & 2.1040e+00 (1.8214e-01)~$\mathbf{-}$ & 1.3733e+00 (3.3794e-01)~$\mathbf{-}$ & \textbf{9.6333e-02 (1.8901e-02)}~$\mathbf{\approx}$ & 1.7585e-01 (1.2440e-01)~$\mathbf{-}$          & 9.7543e-02 (3.9146e-02)          \\ \cline{2-7} 
                              & 25\%     & 2.9738e+00 (1.8514e-01)~$\mathbf{-}$ & 2.3535e+00 (2.1429e-01)~$\mathbf{-}$ & 2.0568e-01 (3.5956e-02)~$\mathbf{\approx}$          & 5.2164e-01 (5.8015e-01)~$\mathbf{-}$          & \textbf{2.0331e-01 (7.4606e-02)} \\ \cline{2-7} 
                              & 50\%     & 3.5819e+00 (1.6709e-01)~$\mathbf{-}$ & 2.9171e+00 (2.5604e-01)~$\mathbf{-}$ & 3.5220e-01 (5.6119e-02)~$\mathbf{\approx}$          & 7.7263e-01 (7.7222e-01)~$\mathbf{-}$          & \textbf{3.2185e-01 (1.1253e-01)} \\ \cline{2-7} 
                              & 75\%     & 4.5056e+00 (1.1115e-01)~$\mathbf{-}$ & 3.6170e+00 (2.8501e-01)~$\mathbf{-}$ & 6.7452e-01 (1.0479e-01)~$\mathbf{-}$          & 9.9810e-01 (7.4903e-01)~$\mathbf{-}$          & \textbf{4.8967e-01 (1.7913e-01)} \\ \cline{2-7} 
                              & 95\%     & 6.0673e+00 (2.1242e-01)~$\mathbf{-}$ & 5.0808e+00 (1.6938e-01)~$\mathbf{-}$ & 2.0580e+00 (1.5660e-02)~$\mathbf{-}$          & 1.9350e+00 (1.6146e+00)~$\mathbf{-}$          & \textbf{9.1072e-01 (4.9060e-01)} \\ \hline\hline
\multirow{5}{*}{Ackley-II}    & 5\%      & 3.6226e+00 (1.7790e-01)~$\mathbf{-}$ & 3.7484e+00 (2.4236e-02)~$\mathbf{-}$ & 3.7061e+00 (4.6585e-02)~$\mathbf{-}$          & 3.4072e+00 (2.8914e-01)~$\mathbf{-}$          & \textbf{2.5089e+00 (7.6250e-01)} \\ \cline{2-7} 
                              & 25\%     & 4.3273e+00 (1.9539e-01)~$\mathbf{-}$ & 4.3649e+00 (1.6628e-02)~$\mathbf{-}$ & 4.2669e+00 (6.6596e-02)~$\mathbf{-}$          & 4.0061e+00 (1.9874e-01)~$\mathbf{\approx}$          & \textbf{3.5385e+00 (7.6554e-01)} \\ \cline{2-7} 
                              & 50\%     & 4.9140e+00 (2.6313e-01)~$\mathbf{\approx}$ & 4.8317e+00 (1.2213e-01)~$\mathbf{\approx}$ & 4.6222e+00 (6.0964e-02)~$\mathbf{\approx}$          & 4.4228e+00 (1.6292e-01)~$\mathbf{\approx}$          & \textbf{4.3123e+00 (8.1509e-01)} \\ \cline{2-7} 
                              & 75\%     & 5.5161e+00 (2.7997e-01)~$\mathbf{\approx}$ & 5.4248e+00 (1.3343e-01)~$\mathbf{\approx}$ & 5.2040e+00 (1.5259e-01)~$\mathbf{\approx}$          & 4.9905e+00 (2.3661e-01)~$\mathbf{\approx}$          & \textbf{4.9644e+00 (8.1048e-01)} \\ \cline{2-7} 
                              & 95\%     & 6.4260e+00 (4.2145e-01)~$\mathbf{\approx}$ & 6.0830e+00 (1.8482e-01)~$\mathbf{\approx}$ & 5.8336e+00 (1.3509e-01)~$\mathbf{\approx}$          & \textbf{5.6838e+00 (2.5812e-01)}~$\mathbf{\approx}$ & 5.8641e+00 (7.5154e-01)          \\ \hline\hline
\multirow{5}{*}{Rastrigin-I}  & 5\%      & 2.6359e+01 (6.2784e+00)~$\mathbf{\approx}$  & 3.6730e+01 (7.6639e+00)~$\mathbf{-}$ & \textbf{2.5169e+01 (2.6266e+00)}~$\mathbf{\approx}$  & 2.7618e+01 (3.6902e+00)~$\mathbf{\approx}$          & 2.6238e+01 (2.5165e+00)         \\ \cline{2-7} 
                              & 25\%     & 4.4250e+01 (7.7101e+00)~$\mathbf{\approx}$  & 5.7113e+01 (1.2497e+01)~$\mathbf{-}$ & \textbf{4.2094e+01 (3.1510e+00)}~$\mathbf{\approx}$  & 4.4928e+01 (5.5935e+00)~$\mathbf{\approx}$          & 4.2326e+01 (3.6744e+00)          \\ \cline{2-7} 
                              & 50\%     & 5.9753e+01 (1.0151e+01)~$\mathbf{-}$  & 7.4654e+01 (1.8622e+01)~$\mathbf{-}$ & 5.6112e+01 (3.9190e+00)~$\mathbf{\approx}$           & 5.9310e+01 (8.8632e+00)~$\mathbf{\approx}$          & \textbf{5.4785e+01 (5.4205e+00)} \\ \cline{2-7} 
                              & 75\%     & 8.3515e+01 (1.8001e+01)~$\mathbf{-}$  & 9.8313e+01 (2.8369e+01)~$\mathbf{-}$ & 7.4887e+01 (6.2412e+00)~$\mathbf{-}$           & 7.9073e+01 (1.9286e+01)~$\mathbf{-}$         & \textbf{6.8098e+01 (8.2728e+00)} \\ \cline{2-7} 
                              & 95\%     & 1.5818e+02 (4.3552e+01)~$\mathbf{-}$  & 1.6669e+02 (4.0435e+01)~$\mathbf{-}$ & 1.3682e+02 (2.2979e+01)~$\mathbf{-}$           & 1.2808e+02 (5.8301e+01)~$\mathbf{-}$          & \textbf{8.9259e+01 (1.5389e+01)} \\ \hline\hline
\multirow{5}{*}{Rastrigin-II} & 5\%      & 6.4737e+01 (1.7582e+00)~$\mathbf{-}$ & 6.7883e+01 (1.5605e+00)~$\mathbf{-}$ & 6.3057e+01 (2.2005e+00)~$\mathbf{-}$          & 6.1093e+01 (2.8075e+00)~$\mathbf{-}$          & \textbf{4.5054e+01 (6.0348e+00)} \\ \cline{2-7} 
                              & 25\%     & 9.7163e+01 (3.5574e+00)~$\mathbf{-}$ & 9.4716e+01 (1.7024e+00)~$\mathbf{-}$ & 9.3105e+01 (2.6550e+00)~$\mathbf{-}$          & 9.6881e+01 (4.1387e+00)~$\mathbf{-}$          & \textbf{7.1361e+01 (1.1557e+01)} \\ \cline{2-7} 
                              & 50\%     & 1.2564e+02 (6.9602e+00)~$\mathbf{-}$ & 1.1717e+02 (2.2589e+00)~$\mathbf{-}$ & 1.1905e+02 (3.9188e+00)~$\mathbf{-}$          & 1.2988e+02 (9.3123e+00)~$\mathbf{-}$          & \textbf{9.4798e+01 (1.6858e+01)} \\ \cline{2-7} 
                              & 75\%     & 1.6150e+02 (1.3363e+01)~$\mathbf{-}$ & 1.4235e+02 (2.8370e+00)~$\mathbf{-}$ & 1.5109e+02 (6.5386e+00)~$\mathbf{-}$          & 1.6934e+02 (1.5657e+01)~$\mathbf{-}$          & \textbf{1.2334e+02 (2.3236e+01)} \\ \cline{2-7} 
                              & 95\%     & 2.2799e+02 (2.9678e+01)~$\mathbf{-}$ & 1.8137e+02 (5.4616e+00)~$\mathbf{\approx}$ & 2.0929e+02 (1.4155e+01)~$\mathbf{\approx}$          & 2.3501e+02 (2.6453e+01)~$\mathbf{-}$          & \textbf{1.7181e+02 (3.4438e+01)} \\ \hline\hline
\multirow{5}{*}{Griewank-I}   & 5\%      & 1.4295e+00 (3.6791e-01)~$\mathbf{\approx}$ & 1.6501e+00 (2.4926e-01)~$\mathbf{-}$ & \textbf{1.2337e+00 (1.7745e-01)}~$\mathbf{\approx}$ & 1.5408e+00(2.4038e-01)~$\mathbf{-}$           & 1.3317e+00 (1.5637e-01)          \\ \cline{2-7} 
                              & 25\%     & 2.9319e+00 (1.5217e+00)~$\mathbf{-}$ & 3.3922e+00 (9.2390e-01)~$\mathbf{-}$ & 2.1563e+00 (6.2049e-01)~$\mathbf{\approx}$          & 2.6166e+00(5.5361e-01)~$\mathbf{-}$           & \textbf{2.0979e+00 (4.5880e-01)} \\ \cline{2-7} 
                              & 50\%     & 5.6758e+00 (3.7501e+00)~$\mathbf{-}$ & 5.6437e+00 (1.6059e+00)~$\mathbf{-}$ & 3.7437e+00 (1.3372e+00)~$\mathbf{\approx}$          & 4.1832e+00(1.1701e+00)~$\mathbf{-}$           & \textbf{3.0935e+00 (1.0398e+00)} \\ \cline{2-7} 
                              & 75\%     & 1.1592e+01 (7.4868e+00)~$\mathbf{-}$ & 1.0284e+01 (2.4744e+00)~$\mathbf{-}$ & 7.8351e+00 (2.3734e+00)~$\mathbf{-}$          & 7.8516e+00(3.4542e+00)~$\mathbf{-}$           & \textbf{4.7885e+00 (1.9504e+00)} \\ \cline{2-7} 
                              & 95\%     & 2.8449e+01 (1.2148e+01)~$\mathbf{-}$ & 2.7646e+01 (5.1187e+00)~$\mathbf{-}$ & 2.4819e+01 (5.0000e+00)~$\mathbf{-}$          & 1.9363e+01(1.1093e+01)~$\mathbf{-}$           & \textbf{8.5325e+00 (4.3527e+00)} \\ \hline\hline
\multirow{5}{*}{Griewank-II}  & 5\%      & 8.5584e+00 (4.8922e-01)~$\mathbf{-}$ & 8.7020e+00 (3.1887e-01)~$\mathbf{-}$ & 8.1302e+00 (5.1623e-01)~$\mathbf{-}$          & 6.7492e+00 (1.2233e+00)~$\mathbf{-}$          & \textbf{4.0665e+00 (1.3843e+00)} \\ \cline{2-7} 
                              & 25\%     & 1.3381e+01 (6.6195e-01)~$\mathbf{-}$ & 1.3495e+01 (1.0838e-01)~$\mathbf{-}$ & 1.3655e+01 (1.6315e-01)~$\mathbf{-}$          & 1.2651e+01 (1.6883e+00)~$\mathbf{-}$          & \textbf{9.0974e+00 (3.1722e+00)} \\ \cline{2-7} 
                              & 50\%     & 1.8848e+01 (1.3946e+00)~$\mathbf{-}$ & 1.8258e+01 (3.7859e-01)~$\mathbf{\approx}$ & 1.9214e+01 (5.9426e-01)~$\mathbf{-}$          & 1.8282e+01 (2.5765e+00)~$\mathbf{\approx}$          & \textbf{1.4656e+01 (4.2747e+00)} \\ \cline{2-7} 
                              & 75\%     & 2.5659e+01 (2.6930e+00)~$\mathbf{\approx}$ & 2.3927e+01 (8.2098e-01)~$\mathbf{\approx}$ & 2.6220e+01 (8.5712e-01)~$\mathbf{\approx}$          & 2.5084e+01 (3.7183e+00)~$\mathbf{\approx}$          & \textbf{2.2094e+01 (5.0576e+00)} \\ \cline{2-7} 
                              & 95\%     & 3.7448e+01 (6.2271e+00)~$\mathbf{\approx}$ & \textbf{3.2592e+01 (1.6257e+00)}~$\mathbf{\approx}$ & 3.8192e+01 (1.5611e+00)~$\mathbf{\approx}$          & 3.7894e+01 (6.7145e+00)~$\mathbf{\approx}$          & 3.5562e+01 (6.5583e+00) \\ \hline\hline
\end{tabular}
}
\label{tab:synthetic}
\end{table*}

Fig.~\ref{Fig: combined-convergence} presents the offline optimization performance of the considered methods on sample tasks from two representative synthetic test problems. 
It can be observed that PMTO-FT achieves significantly better empirical convergence than the single-task baseline in the offline optimization stage.
This improvement verifies the theory discussed in Section IV and in the supplementary material.
Moreover, SELF, the expensive MTO method can outperform the single-task baseline but is generally inferior compared with PMTO variants in the offline optimization stage, as shown in Fig.\ref{Fig: combined-convergence}.
The improvement over the single-task counterpart can be attributed to the knowledge transfer capabilities of multi-task GP models in SELF~\cite{SELF} and the inferior results compared to PMTO variants may be due to the lack of full utilization of task-specific side information in the optimization loop.
Crucially, these high-quality solutions from offline optimization are essential for training an effective task model's online performance, as shown in TABLE \ref{tab:synthetic}.

TABLE \ref{tab:synthetic} summarizes the task model's online performance across all sampled task parameters for each quantile, comparing GP-based optimization (single-task baseline method), SELF (MTO baseline method), PMTO-FT (PMTO without task evolution module), $(\boldsymbol{\theta}_l, \boldsymbol{\theta}_u)$-PMTO-RT (PMTO replacing task evolution with random sampling) and $(\boldsymbol{\theta}_l, \boldsymbol{\theta}_u)$-PMTO.
Notably, $(\boldsymbol{\theta}_l, \boldsymbol{\theta}_u)$-PMTO outperforms the other methods in 31 out of 40 quantiles, highlighting its ability to achieve better online optimized results across the incoming task parameters. 
Moreover, PMTO-FT surpasses its single-task counterpart, GP-based optimization, in 32 of 40 quantiles, demonstrating the benefits of attaining better offline optimization results via the utilization of task-specific side information, as shown in Fig.~\ref{Fig: combined-convergence}.
A similar trend can be observed in TABLE \ref{tab:synthetic} where the proposed $(\boldsymbol{\theta}_l, \boldsymbol{\theta}_u)$-PMTO can significantly outperform SELF in 29 out of 40 quantiles.
TABLE \ref{tab:synthetic} also includes comparative results that substantiate the effectiveness of the task evolution module of the proposed $(\boldsymbol{\theta}_l, \boldsymbol{\theta}_u)$-PMTO.
PMTO-FT is a PMTO variant without the task evolution module and $(\boldsymbol{\theta}_l, \boldsymbol{\theta}_u)$-PMTO-RT is a PMTO variant that replaces the strategic task evolution module with randomly sampling a task parameter per iteration. 
Serving as an ablation study, our proposed method significantly outperforms the random search variant and PMTO-FT in most quantiles, underscoring the effectiveness of task evolution in actively exploring the task space during the offline optimization run.

While our task evolution module enables strategic placement of sampled tasks and hence improves the task model's online optimization performance, we observe that in certain low-complexity problems (e.g., Sphere-I, Rastrigin-I), the fixed-task variant PMTO-FT can be comparable or even slightly better than  $(\boldsymbol{\theta}_l, \boldsymbol{\theta}_u)$-PMTO at low quantiles (e.g., 5\%). 
This occurs because diversity-based sampling may occasionally introduce tasks that are distant from the existing task set, leading to reduced refinement of model capabilities in regions of primary interest.
This trade-off between the overall task model performance and the performance at low quantiles with less complex problem sets points to a possible direction for future work to develop finer-grained or multiobjective sampling strategies that balance diversity with task relevance—thus enabling more stable and robust task model construction across different problem regimes.

\begin{table*}
\centering
\caption{Comparative Results of the Task Model's Online Performance on Adaptive Control Systems. $U=20$ independent trials are considered.}
\resizebox{12.5cm}{!}{ 
\begin{tabular}{c|c|c|c|c}
\hline
Problems                       & Quantile & GP-based Optimization   & PMTO-FT                 & $(\boldsymbol{\theta}_l, \boldsymbol{\theta}_u)$-PMTO                             \\ \hline\hline
\multirow{5}{*}{Robot Arm}     & 5\%      & 4.2821e-02 (1.9260e-02)~$\mathbf{-}$ & 3.4766e-02 (7.9264e-03)~$\mathbf{-}$ & \textbf{1.7336e-02 (5.2244e-03)} \\ \cline{2-5} 
                               & 25\%     & 8.9019e-02 (2.5297e-02)~$\mathbf{-}$ & 7.6430e-02 (1.7672e-02)~$\mathbf{-}$ & \textbf{4.7606e-02 (1.4060e-02)} \\ \cline{2-5} 
                               & 50\%     & 1.3401e-01 (2.0803e-02)~$\mathbf{-}$ & 1.0522e-01 (2.0310e-02)~$\mathbf{-}$ & \textbf{7.5639e-02 (1.6956e-02)} \\ \cline{2-5} 
                               & 75\%     & 1.7645e-01 (1.5028e-02)~$\mathbf{-}$ & 1.3607e-01 (1.9514e-02)~$\mathbf{\approx}$ & \textbf{1.1816e-01 (1.8595e-02)} \\ \cline{2-5} 
                               & 95\%     & 2.3556e-01 (1.6606e-02)~$\mathbf{-}$ & 1.7845e-01 (1.4280e-02)~$\mathbf{\approx}$ & \textbf{1.7742e-01 (8.5307e-03)} \\ \hline\hline
\multirow{5}{*}{Crane Load-I}  & 5\%      & 3.0679e+04 (5.7316e+04)~$\mathbf{-}$ & 1.4233e+03 (1.3452e+03)~$\mathbf{-}$ & \textbf{4.6175e+02 (3.2924e+02)} \\ \cline{2-5} 
                               & 25\%     & 1.1283e+05 (1.4729e+05)~$\mathbf{-}$ & 2.7508e+04 (2.0614e+04)~$\mathbf{-}$ & \textbf{1.0081e+04 (4.3555e+03)} \\ \cline{2-5} 
                               & 50\%     & 2.1993e+05 (2.1653e+05)~$\mathbf{-}$ & 1.0485e+05 (7.1414e+04)~$\mathbf{-}$ & \textbf{3.9601e+04 (1.2591e+04)} \\ \cline{2-5} 
                               & 75\%     & 3.7102e+05 (2.9138e+05)~$\mathbf{-}$ & 2.6922e+05 (1.6376e+05)~$\mathbf{-}$ & \textbf{1.4146e+05 (1.2170e+05)} \\ \cline{2-5} 
                               & 95\%     & 6.0364e+05 (3.7526e+05)~$\mathbf{\approx}$ & 6.1116e+05 (3.8202e+05)~$\mathbf{\approx}$ & \textbf{3.4578e+05 (3.2520e+05)} \\ \hline\hline
\multirow{5}{*}{Crane Load-II} & 5\%      & 6.3392e+02 (3.5072e+02)~$\mathbf{-}$ & 4.7310e+02 (1.1462e+02)~$\mathbf{-}$ & \textbf{1.5026e+02 (7.6382e+01)} \\ \cline{2-5} 
                               & 25\%     & 1.6577e+04 (9.7445e+03)~$\mathbf{-}$ & 1.1846e+04 (3.3891e+03)~$\mathbf{-}$ & \textbf{3.0218e+03 (9.1541e+02)} \\ \cline{2-5} 
                               & 50\%     & 7.4319e+04 (4.7370e+04)~$\mathbf{-}$ & 5.7356e+04 (1.8227e+04)~$\mathbf{-}$ & \textbf{1.1835e+04 (2.1282e+03)} \\ \cline{2-5} 
                               & 75\%     & 1.7974e+05 (9.7858e+04)~$\mathbf{-}$ & 1.6482e+05 (5.2057e+04)~$\mathbf{-}$ & \textbf{2.9978e+04 (7.5976e+03)} \\ \cline{2-5} 
                               & 95\%     & 3.9408e+05 (2.0022e+05)~$\mathbf{-}$ & 3.8901e+05 (1.1719e+05)~$\mathbf{-}$ & \textbf{8.8104e+04 (3.5100e+04)} \\ \hline\hline
\end{tabular}}
\label{tab:control}
\end{table*}

\begin{figure*}[!t]
\centering
\subfloat[]{\includegraphics[width=3.6in]{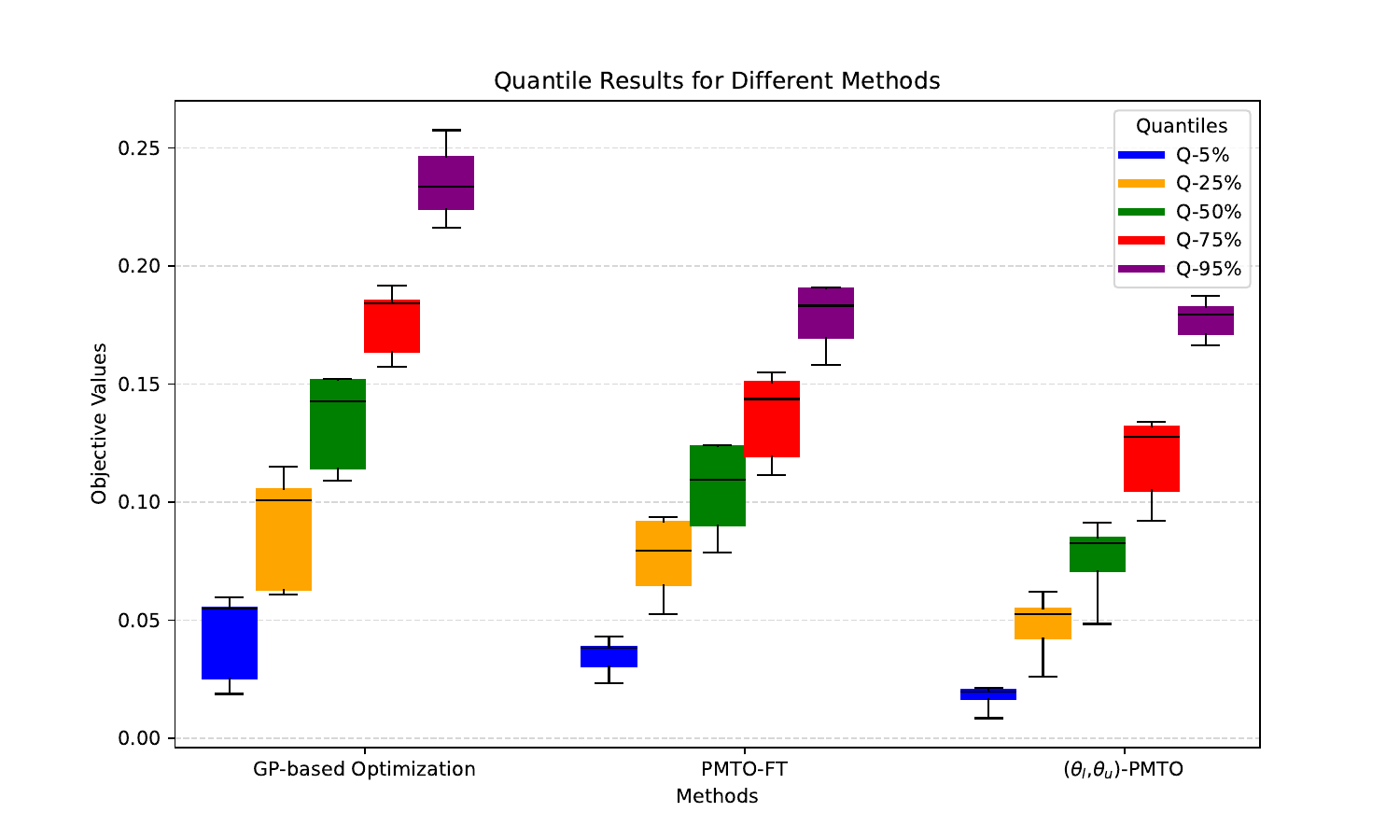}%
\label{fig: exp_sep_arm}}
\subfloat[]{\includegraphics[width=3.6in]{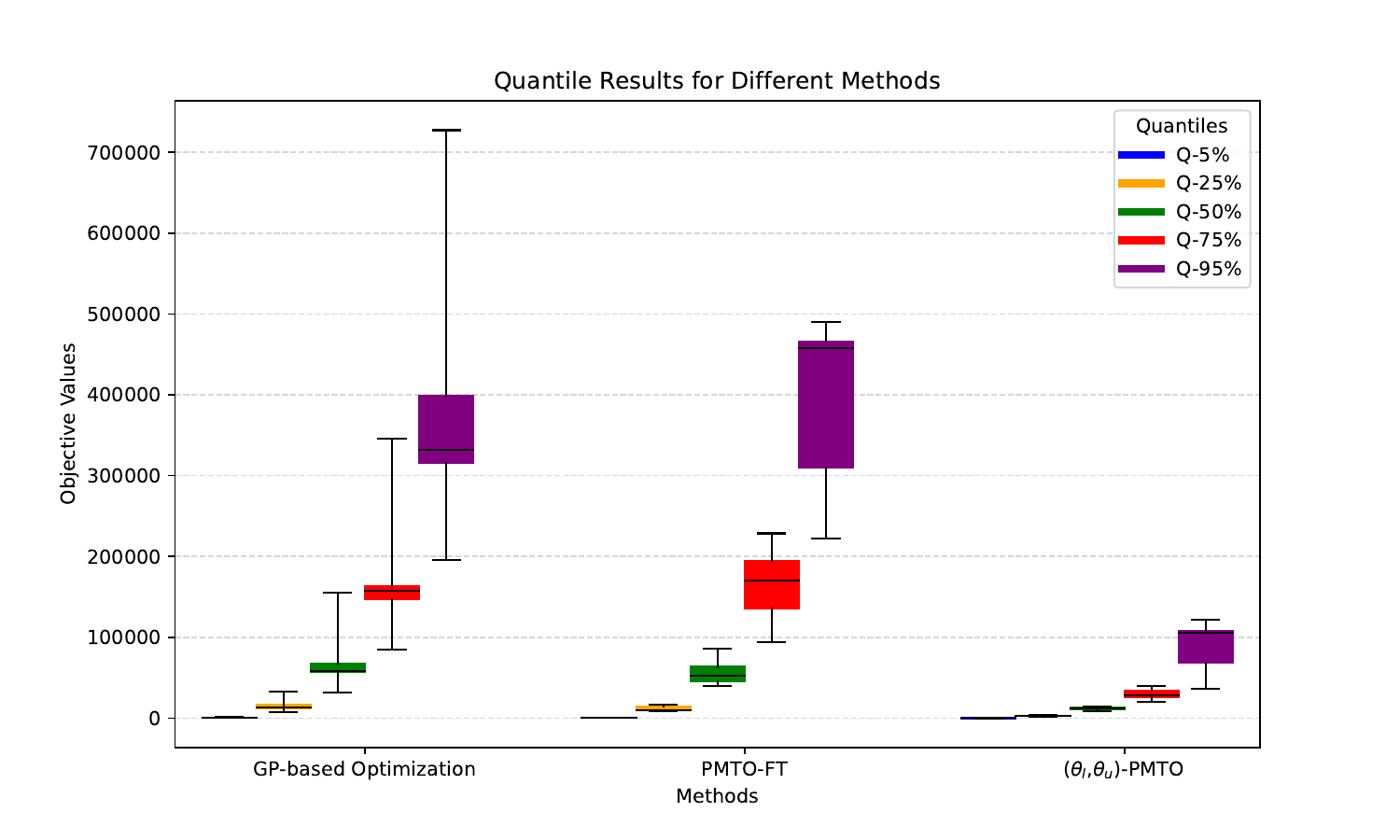}%
\label{fig: exp_recontrol}}
\hfil
\caption{Comparative results of the task model's online performance. The results include single-task baseline method, PMTO-FT, and our proposed PMTO method on adaptive control problems. $U=20$ independent trials are considered: (a) Robot arm optimization under distinct operating conditions (b) Crane-load system optimization under distinct environments.}
\label{fig: exp}
\end{figure*}

\subsection{Case Study in Adaptive Control System}
\subsubsection{Robot Arm Optimization Problem}
As shown in Fig.~3(a), the robot arm optimization problem is to adjust the angular positions of the joints to bring the end effector as close as possible to the target position\footnote{One can refer to \cite{qd-multitask} for the exact formulation of the objective function.}.
Here, the distance between the end effector and the target position is considered the objective function, while the angular positions of the joints serve as the solution~(i.e., $\alpha_1$, $\alpha_2$, $\alpha_3$ in Fig.~3(a)).
The task space includes task parameter $L$ bounded by $[0.5/n, 1/n]$ and $\alpha_{max}$ bounded by $[0.5\pi/n, \pi/n]$, where $L$ is the length of each arm, $\alpha_{max}$ identifies the maximal range of the rotation of each joint, and $n$ is the number of joints set to $n=3$ in our case study.
The position of the first joint attached to the ground is set as [0, 0] while the target position is fixed to [0.5, 0.5] for all optimization tasks.
As shown in TABLE \ref{tab:control} and Fig.~4(a), compared to GP-based optimization, PMTO-FT can stably achieve better optimization performance for the 50\%, 75\%, 95\% quantile optimization results, whereas for the best cases 5\% and 25\%, both optimization methods in the fixed set of tasks do not show significant difference.
The fixed-task methods are limited by constrained task environments, hindering optimization results. 
In contrast, by enabling the joint search in the solution space and the task space, the proposed $(\boldsymbol{\theta}_l, \boldsymbol{\theta}_u)$-PMTO leverages related tasks to enhance convergence and achieve better objective values in best case 5\% optimization results and enhance the moderately worse case optimization results (25\% and 50\%) as well.

\subsubsection{Crane-Load System Optimization with Time Delays~(Crane Load-I)}
As shown in Fig.~3(b), we consider optimizing a crane-load system control problem~\cite{crane}.
The goal is to accelerate the system with crane $m_1$ and load $m_2$ by an external drive force $F$ to achieve a target velocity with minimal time and oscillation.
The solutions for this control system are shown in Fig.~3(b), where $t_1, t_2, t_3$ are decision variables to switch the drive force $F$ during the control process.
In actual operating conditions, however, time delays in the control system can arise from environmental uncertainties and they can destabilize the entire system and induce oscillations~\cite{qiuxin}.
Therefore, the task space includes the possible time delays $\Delta t_1, \Delta t_2, \Delta t_3$ for each time interval.
As shown in TABLE \ref{tab:control}, under various time delays, PMTO-FT can achieve a more stable system control compared to the GP-based optimization, exhibiting the power of multi-task optimization enabled by the inclusion of task parameters in the optimization loop.
Moreover, with the additional capability of the joint search in both solution and task spaces, optimization results in different quantiles can be significantly enhanced by the proposed $(\boldsymbol{\theta}_l, \boldsymbol{\theta}_u)$-PMTO.

\subsubsection{Crane-Load System Optimization with Diverse Operating Conditions~(Crane Load-II)}
Here, we still consider optimizing the control problem for the crane-load system in Fig.~3(b) to accelerate the system with crane $m_1$ and load $m_2$ by an external drive force $F$ to achieve a target velocity with minimal time and oscillation.
Differently, we introduce a distinct task space including the diverse operating conditions.
Diverse types of suspension $l$, loads $m_2$, and resistance $W$ for a fixed crane can occur due to distinct operating conditions and environments.
Therefore, we consider identifying the optimal solutions for diverse possible tuples of ($l$, $m_2$, $W$) in a bounded continuous task space.
Although in a different context, optimization results under distinct environment factors can still be enhanced by PMTO-FT, attributing the merits to the synergies across the system optimization under various operating conditions.
Empowered by \textit{task evolution}, as depicted in TABLE \ref{tab:control} and Fig.~4(b), the joint search across solution and task spaces can further enhance the optimization results across all the quantiles, showing $(\boldsymbol{\theta}_l, \boldsymbol{\theta}_u)$-PMTO's ability to stabilize the crane-load system under diverse operating conditions.

\subsection{Case Study in Robust Engineering Design}
In this case study, we address a robust engineering design problem as shown in Fig.~3(c), where the objective is to minimize the aggregated function of both the structural weight and joint displacement of the truss by optimizing the cross-sectional areas of two bars $\theta_1$, $\theta_2$, and the vertical distance of the second bar $\theta_3$. 
However, due to the existing processing errors, even when these measurements of bars are designed as $\boldsymbol{\theta}=(\theta_1, \theta_2, \theta_3)$, the operating truss structure obtained after actual processing may deviate from the original design.
Let this processing error be represented by $\mathbf{x}=(x_1, x_2, x_3)$ so that the operating structural parameters are ($\boldsymbol{\theta}+\mathbf{x}$).
In robust optimization, the design problem we aim to solve can then be formulated as:
\begin{equation}
    \min_{\boldsymbol{\theta}}\max_{\mathbf{x}}f(\mathbf{x}, \boldsymbol{\theta}).
    \label{eq: minimax}
\end{equation}
Consequently, we show that PMTO can be an effective approach to solve this minimax optimization problem.
We reformulate the minimax problem (\ref{eq: minimax}) by treating the original design variables $\boldsymbol{\theta}$ as task parameters and the processing error $\mathbf{x}$ as the decision vector.
As per the formulation (\ref{eq: task_model}), a task model can be constructed in PMTO to estimate the worst-case processing error corresponding to each design variable by $\mathbf{x}^* = \mathcal{M}(\boldsymbol{\theta})$. 
This obtained task model supports to solve the problem (\ref{eq: minimax}) as $\min_{\boldsymbol{\theta}}f(\mathcal{M}(\boldsymbol{\theta}), \boldsymbol{\theta})$.
Thereafter, the minimax optimization shown as (\ref{eq: minimax}) can be solved by the PMTO.
\begin{figure}
\centering
\includegraphics[width=3in]{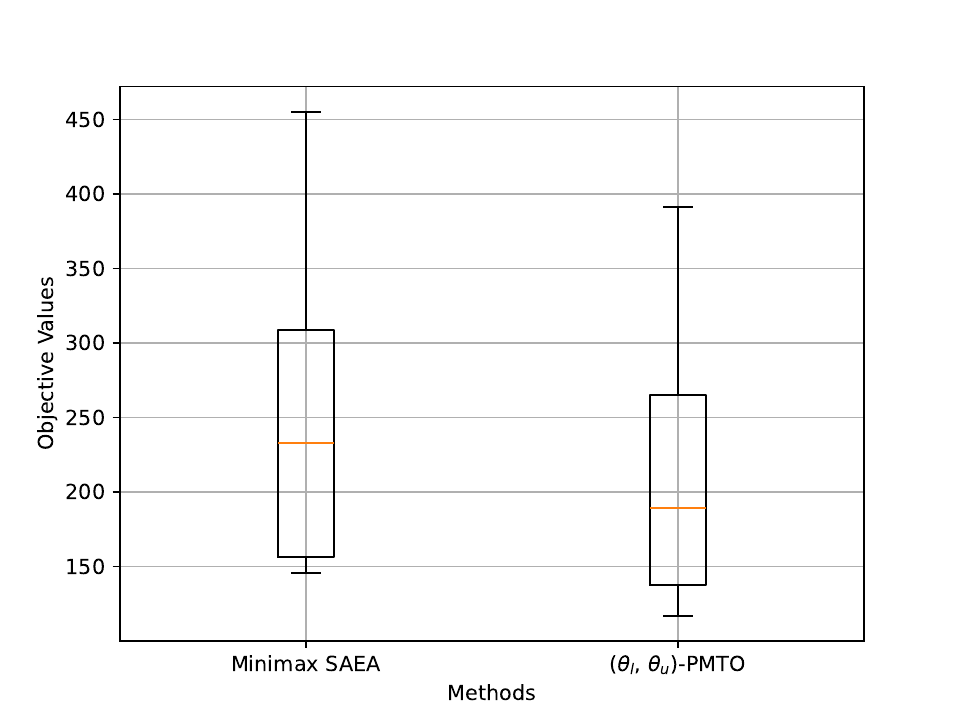}%
\caption{Comparison of design performance under varying processing errors for the minimax optimization problem. The figure shows the objective values of robust design $\boldsymbol{\theta}$ optimized by Minimax SAEA and our proposed $(\boldsymbol{\theta}_l, \boldsymbol{\theta}_u)$-PMTO algorithm, evaluated across diverse processing error $\mathbf{x}$. To ensure a fair comparison, both methods are provided with the same total evaluation budget and assessed on the same set of processing errors.}
\label{fig: minimax}
\end{figure}

As shown in Fig.~\ref{fig: minimax}, we compare $(\boldsymbol{\theta}_l, \boldsymbol{\theta}_u)$-PMTO to a popular minimax optimization solver, Minimax SAEA~\cite{zhou_saea}.
To evaluate the robustness of the generated design, we impose many possible processing errors ($800$ randomly processing errors) on the generated design from both Mimimax SAEA and $(\boldsymbol{\theta}_l, \boldsymbol{\theta}_u)$-PMTO.
As shown in Fig.~\ref{fig: minimax}, $(\boldsymbol{\theta}_l, \boldsymbol{\theta}_u)$-PMTO can generate a relatively more robust plain truss design, showing that the proposed $(\boldsymbol{\theta}_l, \boldsymbol{\theta}_u)$-PMTO can serve as a generic tool to solve minimax optimization as well.

\section{Conclusion}
This paper introduces PMTO as a novel generalization of MTO from a fixed and finite set of tasks to infinite task sets. 
The proposed $(\boldsymbol{\theta}_l, \boldsymbol{\theta}_u)$-PMTO algorithm enables joint exploration of continuous solution and task spaces through two key approximations: mapping the solution space to the objective space for inter-task transfer and mapping the task space to the solution space to target under-explored regions.
Experimental results demonstrate two main benefits.
First, incorporating continuous task parameters as a medium for knowledge transfer accelerates multi-task optimization and improves convergence.
Second, an evolutionary exploration empowered by a calibrated task model in the joint solution-task space enhances the use of information from under-explored areas, boosting overall optimization efficiency.

Despite these advancements, our study primarily addresses low- to moderate-dimensional problems, reflecting the inherent limitations of GP models in high-dimensional search spaces. 
Enhancing the scalability of PMTO methods to extend their applicability to higher-dimensional optimization problems remains a crucial direction for future work. 
Additionally, the application of PMTO to multi-objective optimization settings remains an open problem and represents another promising avenue for future exploration.

\bibliographystyle{IEEEtran}
\bibliography{bibfiles_async}

@article{real-bench,
title = {An easy-to-use real-world multi-objective optimization problem suite},
journal = {Applied Soft Computing},
volume = {89},
pages = {106078},
year = {2020},
issn = {1568-4946},
doi = {https://doi.org/10.1016/j.asoc.2020.106078},
author = {Ryoji Tanabe and Hisao Ishibuchi},
}

@article{da2,
  title={Non-Linear Domain Adaptation in Transfer Evolutionary Optimization},
  author={Lim, Ray and Gupta, Abhishek and Ong, Yew-Soon and Feng, Liang and Zhang, Allan N},
  journal={Cognitive Comput.},
  volume={13},
  pages={290-307},
  year={2021}
}

@inproceedings{effective-diversity,
 author = {Parker-Holder, Jack and Pacchiano, Aldo and Choromanski, Krzysztof M and Roberts, Stephen J},
 booktitle = {Advances in Neural Information Processing Systems},
 pages = {18050--18062},
 publisher = {Curran Associates, Inc.},
 title = {Effective Diversity in Population Based Reinforcement Learning},
 volume = {33},
 year = {2020}
}

@article{dpp,
year = {2012},
volume = {5},
journal = {Foundations and Trends® in Machine Learning},
title = {Determinantal Point Processes for Machine Learning},
doi = {10.1561/2200000044},
issn = {1935-8237},
number = {2–3},
pages = {123-286},
author = {Alex Kulesza and Ben Taskar}
}

@article{GPML,
  title={Gaussian processes for machine learning},
  author={Seeger, Matthias},
  journal={International journal of neural systems},
  volume={14},
  number={02},
  pages={69--106},
  year={2004},
  publisher={World Scientific}
}

@inproceedings{MTGP,
 author = {Bonilla, Edwin V and Chai, Kian and Williams, Christopher},
 booktitle = {Advances in Neural Information Processing Systems},
 editor = {J. Platt and D. Koller and Y. Singer and S. Roweis},
 pages = {},
 publisher = {Curran Associates, Inc.},
 title = {Multi-task {Gaussian} Process Prediction},
 volume = {20},
 year = {2007}
}

@inproceedings{MTBO,
 author = {Swersky, Kevin and Snoek, Jasper and Adams, Ryan P},
 booktitle = {Advances in Neural Information Processing Systems},
 editor = {C.J. Burges and L. Bottou and M. Welling and Z. Ghahramani and K.Q. Weinberger},
 pages = {},
 publisher = {Curran Associates, Inc.},
 title = {Multi-Task {Bayesian} Optimization},
 volume = {26},
 year = {2013}
}

@ARTICLE{osaba-rl,
  author={Martinez, Aritz D. and Del Ser, Javier and Osaba, Eneko and Herrera, Francisco},
  journal={IEEE Trans. on Evol. Comput.}, 
  title={Adaptive Multifactorial Evolutionary Optimization for Multitask Reinforcement Learning}, 
  year={2022},
  volume={26},
  number={2},
  pages={233-247},
  doi={10.1109/TEVC.2021.3083362}}

@inproceedings{confidencebo,
 author = {Sessa, Pier Giuseppe and Laforgue, Pierre and Cesa-Bianchi, Nicol\`{o} and Krause, Andreas},
 booktitle = {Advances in Neural Information Processing Systems},
 editor = {A. Oh and T. Naumann and A. Globerson and K. Saenko and M. Hardt and S. Levine},
 pages = {6770--6781},
 publisher = {Curran Associates, Inc.},
 title = {Multitask Learning with No Regret: from Improved Confidence Bounds to Active Learning},
 volume = {36},
 year = {2023}
}

@InProceedings{safebo,
  title = 	 {Towards safe multi-task {B}ayesian optimization},
  author =       {L\"{u}bsen, Jannis and Hespe, Christian and Eichler, Annika},
  booktitle = 	 {Proceedings of the 6th Annual Learning for Dynamics Control Conference},
  pages = 	 {839--851},
  year = 	 {2024},
  volume = 	 {242},
  series = 	 {Proceedings of Machine Learning Research},
  month = 	 {15--17 Jul},
  publisher =    {PMLR},
}

@INPROCEEDINGS{zhou_saea,
  author={Zhou, Aimin and Zhang, Qingfu},
  booktitle={IEEE Congr. on Evol. Comput.}, 
  title={A surrogate-assisted evolutionary algorithm for minimax optimization}, 
  year={2010},
  volume={},
  number={},
  pages={1-7},
  doi={10.1109/CEC.2010.5586122}}

@inproceedings{qd-multitask,
author = {Mouret, Jean-Baptiste and Maguire, Glenn},
title = {Quality diversity for multi-task optimization},
year = {2020},
isbn = {9781450371285},
publisher = {Association for Computing Machinery},
address = {New York, NY, USA},
doi = {10.1145/3377930.3390203},
booktitle = {Proceedings of the 2020 Genetic and Evol. Comput. Conference},
pages = {121–129},
numpages = {9},
location = {Canc\'{u}n, Mexico},
series = {GECCO '20}
}

@ARTICLE{gmm2,
  author={Min, Alan Tan Wei and Gupta, Abhishek and Ong, Yew-Soon},
  journal={IEEE Trans. Autom. Sci. Eng.}, 
  title={Generalizing Transfer {Bayesian} Optimization to Source-Target Heterogeneity}, 
  year={2021},
  volume={18},
  number={4},
  pages={1754-1765},
  doi={10.1109/TASE.2020.3017644}}

@Article{aq-ei,
author={Zhan, Dawei
and Xing, Huanlai},
title={Expected improvement for expensive optimization: a review},
journal={Journal of Global Optimization},
year={2020},
month={Nov},
day={01},
volume={78},
number={3},
pages={507-544},
issn={1573-2916},
doi={10.1007/s10898-020-00923-x},
}

@article{mp-lp,
author = {Barnett, S.},
title = {A Simple Class of Parametric Linear Programming Problems},
journal = {Operations Research},
volume = {16},
number = {6},
pages = {1160-1165},
year = {1968},
doi = {10.1287/opre.16.6.1160},
}

@article{autonomous-steering-control,
author = {Junho Lee and Hyuk-Jun Chang},
title ={Multi-parametric model predictive control for autonomous steering using an electric power steering system},
journal = {Proceedings of the Institution of Mechanical Engineers, Part D: Journal of Automobile Engineering},
volume = {233},
number = {13},
pages = {3391-3402},
year = {2019},
doi = {10.1177/0954407018824773},
}

@ARTICLE{magnetic-endoscope,
  author={Scaglioni, Bruno and Previtera, Luca and Martin, James and Norton, Joseph and Obstein, Keith L. and Valdastri, Pietro},
  journal={IEEE Robotics and Automation Letters}, 
  title={Explicit Model Predictive Control of a Magnetic Flexible Endoscope}, 
  year={2019},
  volume={4},
  number={2},
  pages={716-723},
  doi={10.1109/LRA.2019.2893418}}

@INPROCEEDINGS{current-control,
  author={Jia, Chengyu and Wang, Xudong and Zhou, Kai and Xianqing, Dai},
  booktitle={2019 IEEE 2nd International Conference on Automation, Electronics and Electrical Engineering (AUTEEE)}, 
  title={Sensorless Explicit Model Predictive Control for {IPMSM} Drives}, 
  year={2019},
  volume={},
  number={},
  pages={82-87},
  doi={10.1109/AUTEEE48671.2019.9033424}}

@INPROCEEDINGS{energy-manage-vehicle,
  author={Vadamalu, Raja Sangili and Beidl, Christian},
  booktitle={2016 European Control Conference (ECC)}, 
  title={Explicit {MPC PHEV} energy management using Markov chain based predictor: Development and validation at Engine-In-The-Loop testbed}, 
  year={2016},
  volume={},
  number={},
  pages={453-458},
  doi={10.1109/ECC.2016.7810326}}

@ARTICLE{multitask-robust,
  author={Wang, Handing and Feng, Liang and Jin, Yaochu and Doherty, John},
  journal={IEEE Comput. Intell. Magazine}, 
  title={Surrogate-Assisted Evolutionary Multitasking for Expensive Minimax Optimization in Multiple Scenarios}, 
  year={2021},
  volume={16},
  number={1},
  pages={34-48},
  doi={10.1109/MCI.2020.3039067}}

@ARTICLE{SELF,
  author={Tan, Shenglian and Wang, Yong and Sun, Guangyong and Pang, Tong and Tang, Ke},
  journal={IEEE Trans. on Evol. Comput.}, 
  title={A Surrogate-Assisted Evolutionary Framework for Expensive Multitask Optimization Problems}, 
  year={2024},
  volume={},
  number={},
  doi={10.1109/TEVC.2024.3370937}}

@inproceedings{multitask-qd, 
author = {Anne, Timoth\'{e}e and Mouret, Jean-Baptiste}, 
title = {Parametric-Task {MAP}-{E}lites}, 
year = {2024}, 
isbn = {9798400704949}, 
publisher = {Association for Computing Machinery}, 
address = {New York, NY, USA}, 
doi = {10.1145/3638529.3653993}, 
booktitle = {Proceedings of the Genetic and Evol. Comput. Conference}, 
pages = {68–77}, 
numpages = {10}, 
location = {Melbourne, VIC, Australia}, 
series = {GECCO '24}
}

@InProceedings{moo-finv,
author="Wei, Tingyang
and Liu, Jiao
and Gupta, Abhishek
and Tan, Puay Siew
and Ong, Yew-Soon",
title="Bayesian Forward-Inverse Transfer for Multiobjective Optimization",
booktitle="Parallel Problem Solving from Nature -- PPSN XVIII",
year="2024",
publisher="Springer Nature Switzerland",
address="Cham",
pages="135--152",
isbn="978-3-031-70085-9"
}

@article{mp-mip,
title = {Multi-parametric mixed integer linear programming under global uncertainty},
journal = {Computers \& Chemical Engineering},
volume = {116},
pages = {279-295},
year = {2018},
issn = {0098-1354},
author = {Vassilis M. Charitopoulos and Lazaros G. Papageorgiou and Vivek Dua},
}

@article{mp-mpc,
title = {{MPC} on a chip—Recent advances on the application of multi-parametric model-based control},
journal = {Computers \& Chemical Engineering},
volume = {32},
number = {4},
pages = {754-765},
year = {2008},
issn = {0098-1354},
doi = {https://doi.org/10.1016/j.compchemeng.2007.03.008},
author = {P. Dua and K. Kouramas and V. Dua and E.N. Pistikopoulos},
}

@article{mp-nlp,
title = {On-line optimization via off-line parametric optimization tools},
journal = {Computers \& Chemical Engineering},
volume = {26},
number = {2},
pages = {175-185},
year = {2002},
issn = {0098-1354},
doi = {https://doi.org/10.1016/S0098-1354(01)00739-6},
author = {Efstratios N. Pistikopoulos and Vivek Dua and Nikolaos A. Bozinis and Alberto Bemporad and Manfred Morari},
}

@ARTICLE{mp-survey,
AUTHOR={Pappas, Iosif  and Kenefake, Dustin  and Burnak, Baris  and Avraamidou, Styliani  and Ganesh, Hari S.  and Katz, Justin  and Diangelakis, Nikolaos A.  and Pistikopoulos, Efstratios N. },
TITLE={Multiparametric Programming in Process Systems Engineering: Recent Developments and Path Forward},
JOURNAL={Frontiers in Chemical Engineering},
VOLUME={2},
YEAR={2021},
DOI={10.3389/fceng.2020.620168},
ISSN={2673-2718},
}

@InProceedings{aq-es,
  title = 	 {Max-value Entropy Search for Efficient {B}ayesian Optimization},
  author =       {Zi Wang and Stefanie Jegelka},
  booktitle = 	 {Proceedings of the 34th International Conference on Machine Learning},
  pages = 	 {3627--3635},
  year = 	 {2017},
  volume = 	 {70},
  series = 	 {Proceedings of Machine Learning Research},
  month = 	 {06--11 Aug},
  publisher =    {PMLR},
}

@article{aq-kg,
title = {Continuous multi-task {Bayesian} Optimisation with correlation},
journal = {European Journal of Operational Research},
volume = {270},
number = {3},
pages = {1074-1085},
year = {2018},
issn = {0377-2217},
doi = {https://doi.org/10.1016/j.ejor.2018.03.017},
author = {Michael Pearce and Juergen Branke},
}

@ARTICLE{andrea,
  author={Srinivas, Niranjan and Krause, Andreas and Kakade, Sham M. and Seeger, Matthias W.},
  journal={IEEE Trans. on Information Theory}, 
  title={Information-Theoretic Regret Bounds for {Gaussian} Process Optimization in the Bandit Setting}, 
  year={2012},
  volume={58},
  number={5},
  pages={3250-3265},
  doi={10.1109/TIT.2011.2182033}}

@ARTICLE{human-loop,
  author={Shahriari, Bobak and Swersky, Kevin and Wang, Ziyu and Adams, Ryan P. and de Freitas, Nando},
  journal={Proceedings of the IEEE}, 
  title={Taking the Human Out of the Loop: A Review of {Bayesian} Optimization}, 
  year={2016},
  volume={104},
  number={1},
  pages={148-175},
  doi={10.1109/JPROC.2015.2494218}}

@article{gmfea,
  title={Generalized multitasking for evolutionary optimization of expensive problems},
  author={Ding, Jinliang and Yang, Cuie and Jin, Yaochu and Chai, Tianyou},
  journal={IEEE Trans. on Evol. Comput.},
  volume={23},
  number={1},
  pages={44--58},
  year={2017},
  publisher={IEEE}
}

@ARTICLE{combinatorial-multitask,
  author={Feng, Liang and Huang, Yuxiao and Zhou, Lei and Zhong, Jinghui and Gupta, Abhishek and Tang, Ke and Tan, Kay Chen},
  journal={IEEE Trans. on Cybern.}, 
  title={Explicit Evolutionary Multitasking for Combinatorial Optimization: A Case Study on Capacitated Vehicle Routing Problem}, 
  year={2021},
  volume={51},
  number={6},
  pages={3143-3156},
  doi={10.1109/TCYB.2019.2962865}}

@ARTICLE{yuxiao,
  author={Huang, Yuxiao and Zhou, Wei and Wang, Yu and Li, Min and Feng, Liang and Tan, Kay Chen},
  journal={IEEE Trans. on Evol. Comput.}, 
  title={Evolutionary Multitasking With Centralized Learning for Large-Scale Combinatorial Multiobjective Optimization}, 
  year={2024},
  volume={28},
  number={5},
  pages={1499-1513},
  doi={10.1109/TEVC.2023.3323877}}

@ARTICLE{wei,
  author={Wei, Tingyang and Wang, Shibin and Zhong, Jinghui and Liu, Dong and Zhang, Jun},
  journal={IEEE Trans. on Evol. Comput.}, 
  title={A Review on Evolutionary Multitask Optimization: Trends and Challenges}, 
  year={2022},
  volume={26},
  number={5},
  pages={941-960},
  doi={10.1109/TEVC.2021.3139437}}

@ARTICLE{geodesic-flow,
  author={Tang, Zedong and Gong, Maoguo and Wu, Yue and Qin, A. K. and Tan, Kay Chen},
  journal={IEEE Trans. on Cybern.}, 
  title={A Multifactorial Optimization Framework Based on Adaptive Intertask Coordinate System}, 
  year={2022},
  volume={52},
  number={7},
  pages={6745-6758},
  doi={10.1109/TCYB.2020.3043509}}

@article{gong2,
 author={Tang, Zedong and Gong, Maoguo and Wu, Yue and Liu, Wenfeng and Xie, Yu},
  journal={IEEE Trans. on Evol. Comput.}, 
  title={Regularized Evolutionary Multitask Optimization: Learning to Intertask Transfer in Aligned Subspace}, 
  year={2021},
  volume={25},
  number={2},
  pages={262-276},
  doi={10.1109/TEVC.2020.3023480}}

@inproceedings{lda,
  title={Linearized domain adaptation in evolutionary multitasking},
  author={Bali, Kavitesh Kumar and Gupta, Abhishek and Feng, Liang and Ong, Yew Soon and Siew, Tan Puay},
  booktitle={2017 IEEE Congr. on Evol. Comput. (CEC)},
  pages={1295--1302},
  year={2017},
}

@INPROCEEDINGS{multiform,
  author={Da, Bingshui and Gupta, Abhishek and Ong, Yew-Soon and Feng, Liang},
  booktitle={2016 IEEE Congr. on Evol. Comput. (CEC)}, 
  title={Evolutionary multitasking across single and multi-objective formulations for improved problem solving}, 
  year={2016},
  volume={},
  number={},
  pages={1695-1701},
  doi={10.1109/CEC.2016.7743992}}

@ARTICLE{dozen,
  author={Gupta, Abhishek and Zhou, Lei and Ong, Yew-Soon and Chen, Zefeng and Hou, Yaqing},
  journal={IEEE Comput. Intell. Mag.}, 
  title={Half a Dozen Real-World Applications of Evolutionary Multitasking, and More}, 
  year={2022},
  volume={17},
  number={2},
  pages={49-66},
  doi={10.1109/MCI.2022.3155332}}

@ARTICLE{car_new,
  author={Rios, Thiago and van Stein, Bas and Bäck, Thomas and Sendhoff, Bernhard and Menzel, Stefan},
  journal={IEEE Trans. on Evol. Comput.}, 
  title={Multi-Task Shape Optimization Using a {3D} Point Cloud Autoencoder as Unified Representation}, 
  year={2021},
  volume={},
  number={},
  pages={1-1},
  doi={10.1109/TEVC.2021.3086308}}

@ARTICLE{ong-discuss,
  author={A. {Gupta} and Y. {Ong} and L. {Feng}},
  journal={IEEE Trans. on Emerg. Topics in Comput. Intell.},
  title={Insights on Transfer Optimization: Because Experience is the Best Teacher},
  year={2018},
  volume={2},
  number={1},
  pages={51-64},
  doi={10.1109/TETCI.2017.2769104}}

@article{mfea,
  title={Multifactorial evolution: toward evolutionary multitasking},
  author={Gupta, Abhishek and Ong, Yew-Soon and Feng, Liang},
  journal={IEEE Trans. on Evol. Comput.},
  volume={20},
  number={3},
  pages={343--357},
  year={2016},
  publisher={IEEE}
}

@article{moomfea,
  title={Multiobjective multifactorial optimization in evolutionary multitasking},
  author={Gupta, Abhishek and Ong, Yew-Soon and Feng, Liang and Tan, Kay Chen},
  journal={IEEE Trans. on Cybern.},
  volume={47},
  number={7},
  pages={1652--1665},
  year={2016},
  publisher={IEEE}
}

@article{multi-benchmark,
  title={Evolutionary multitasking for multiobjective continuous optimization: Benchmark problems, performance metrics and baseline results},
  author={Yuan, Yuan and Ong, Yew-Soon and Feng, Liang and Qin, A Kai and Gupta, Abhishek and Da, Bingshui and Zhang, Qingfu and Tan, Kay Chen and Jin, Yaochu and Ishibuchi, Hisao},
  journal={arXiv preprint arXiv:1706.02766},
  year={2017}
}

@article{mfea2,
  title={Multifactorial evolutionary algorithm with online transfer parameter estimation: {MFEA-II}},
  author={Bali, Kavitesh Kumar and Ong, Yew-Soon and Gupta, Abhishek and Tan, Puay Siew},
  journal={IEEE Trans. on Evol. Comput.},
  volume={24},
  number={1},
  pages={69--83},
  year={2019},
  publisher={IEEE}
}

@ARTICLE{domain,
  author={Wang, Xiaoling and Kang, Qi and Zhou, MengChu and Yao, Siya and Abusorrah, Abdullah},
  journal={IEEE Trans. on Cybern.}, 
  title={Domain Adaptation Multitask Optimization}, 
  year={2023},
  volume={53},
  number={7},
  pages={4567-4578},
  doi={10.1109/TCYB.2022.3222101}}

@ARTICLE{sysu,
  author={Z. {Chen} and Y. {Zhou} and X. {He} and J. {Zhang}},
  journal={IEEE Trans. on Cybern.},
  title={Learning Task Relationships in Evolutionary Multitasking for Multiobjective Continuous Optimization},
  year={2020},
  volume={},
  number={},
  pages={1-12},
  doi={10.1109/TCYB.2020.3029176}}

@ARTICLE{mfgp,
  author={J. {Zhong} and L. {Feng} and W. {Cai} and Y. -S. {Ong}},
  journal={IEEE Trans. on Syst., Man, and Cyben.: Syst.},
  title={Multifactorial Genetic Programming for Symbolic Regression Problems},
  year={2020},
  volume={50},
  number={11},
  pages={4492-4505},
  doi={10.1109/TSMC.2018.2853719}}

@article{min,
  title={Multiproblem surrogates: Transfer evolutionary multiobjective optimization of computationally expensive problems},
  author={Min, Alan Tan Wei and Ong, Yew-Soon and Gupta, Abhishek and Goh, Chi-Keong},
  journal={IEEE Trans. on Evol. Comput.},
  volume={23},
  number={1},
  pages={15--28},
  year={2017},
  publisher={IEEE}
}

@article{fuzzy,
  title={Multitasking Genetic Algorithm ({MTGA}) for Fuzzy System Optimization},
  author={Wu, Dongrui and Tan, Xianfeng},
  journal={IEEE Trans. on Fuzzy Syst.},
  volume={28},
  number={6},
  pages={1050--1061},
  year={2020},
  publisher={IEEE}
}

@ARTICLE{liang-1,
  author={Liang, Zhengping and Liang, Weiqi and Wang, Zhiqiang and Ma, Xiaoliang and Liu, Ling and Zhu, Zexuan},
  journal={IEEE Trans. on Syst., Man, and Cybern.: Syst.}, 
  title={Multiobjective Evolutionary Multitasking With Two-Stage Adaptive Knowledge Transfer Based on Population Distribution}, 
  year={2022},
  volume={52},
  number={7},
  pages={4457-4469},
  doi={10.1109/TSMC.2021.3096220}}

@inproceedings{nips-ong,
 author = {Krause, Andreas and Ong, Cheng},
 booktitle = {Advances in Neural Information Processing Systems},
 editor = {J. Shawe-Taylor and R. Zemel and P. Bartlett and F. Pereira and K.Q. Weinberger},
 pages = {},
 publisher = {Curran Associates, Inc.},
 title = {Contextual {Gaussian} Process Bandit Optimization},
 volume = {24},
 year = {2011}
}

@inproceedings{inverse-density,
author = {Yan, Yiming and Giagkiozis, Ioannis and Fleming, Peter J.},
title = {Improved Sampling of Decision Space for Pareto Estimation},
year = {2015},
isbn = {9781450334723},
publisher = {Association for Computing Machinery},
address = {New York, NY, USA},
doi = {10.1145/2739480.2754713},
booktitle = {Proceedings of the 2015 Annual Conference on Genetic and Evolutionary Computation},
pages = {767–774},
numpages = {8},
location = {Madrid, Spain},
series = {GECCO '15}
}

@Article{submodular,
author={Nemhauser, G. L.
and Wolsey, L. A.
and Fisher, M. L.},
title={An analysis of approximations for maximizing submodular set functions---{I}},
journal={Mathematical Programming},
year={1978},
month={Dec},
day={01},
volume={14},
number={1},
pages={265-294},
issn={1436-4646},
doi={10.1007/BF01588971},
}

@inproceedings{dpp-properties,
 author = {Gillenwater, Jennifer and Kulesza, Alex and Taskar, Ben},
 booktitle = {Advances in Neural Information Processing Systems},
 editor = {F. Pereira and C.J. Burges and L. Bottou and K.Q. Weinberger},
 pages = {},
 publisher = {Curran Associates, Inc.},
 title = {Near-Optimal MAP Inference for Determinantal Point Processes},
 volume = {25},
 year = {2012}
}

@inproceedings{dai2020multitask,
  title     = {Multi-task {Bayesian} Optimization via {Gaussian} Process Upper Confidence Bound},
  author    = {Dai, Sihui and Song, Jialin and Yue, Yisong},
  booktitle = {ICML 2020 Workshop on Real World Experiment Design and Active Learning},
  year      = {2020}
}

@inproceedings{meta-cbo,
 author = {Feng , Qing and Letham, Ben and Mao, Hongzi and Bakshy, Eytan},
 booktitle = {Advances in Neural Information Processing Systems},
 pages = {22032--22044},
 publisher = {Curran Associates, Inc.},
 title = {High-Dimensional Contextual Policy Search with Unknown Context Rewards using {Bayesian} Optimization},
 volume = {33},
 year = {2020}
}

@article{kg-base,
author = {Ginsbourger, David and Baccou, Jean and Chevalier, Cl\'{e}ment and Perales, Fr\'{e}d\'{e}ric and Garland, Nicolas and Monerie, Yann},
title = {{Bayesian} Adaptive Reconstruction of Profile Optima and Optimizers},
journal = {SIAM/ASA Journal on Uncertainty Quantification},
volume = {2},
number = {1},
pages = {490-510},
year = {2014},
doi = {10.1137/130949555},
}

@InProceedings{kernel-bandits,
  title = 	 {On Kernelized Multi-armed Bandits},
  author =       {Sayak Ray Chowdhury and Aditya Gopalan},
  booktitle = 	 {Proceedings of the 34th International Conference on Machine Learning},
  pages = 	 {844--853},
  year = 	 {2017},
  editor = 	 {Precup, Doina and Teh, Yee Whye},
  volume = 	 {70},
  series = 	 {Proceedings of Machine Learning Research},
  month = 	 {06--11 Aug},
  publisher =    {PMLR},
}

@inproceedings{offline-contextual,
 author = {Char, Ian and Chung, Youngseog and Neiswanger, Willie and Kandasamy, Kirthevasan and Nelson, Andrew Oakleigh and Boyer, Mark and Kolemen, Egemen and Schneider, Jeff},
 booktitle = {Advances in Neural Information Processing Systems},
 pages = {},
 publisher = {Curran Associates, Inc.},
 title = {Offline Contextual {Bayesian} Optimization},
 volume = {32},
 year = {2019}
}

@article{autoencoding,
  title={Evolutionary multitasking via explicit autoencoding},
  author={Feng, Liang and Zhou, Lei and Zhong, Jinghui and Gupta, Abhishek and Ong, Yew-Soon and Tan, Kay-Chen and Qin, Alex Kai},
  journal={IEEE Trans. on Cybern.},
  volume={49},
  number={9},
  pages={3457--3470},
  year={2018},
  publisher={IEEE}
}

@article{crane, 
author = {Yuriy, Romasevych and Viatcheslav, Loveikin and Borys, Bakay}, 
title = {A Real-World Benchmark Problem for Global Optimization}, 
year = {2023}, 
issue_date = {Sep 2023}, 
publisher = {Walter de Gruyter GmbH}, 
address = {Berlin, DEU}, 
volume = {23}, number = {3}, 
issn = {1314-4081}, 
doi = {10.2478/cait-2023-0022}, 
journal = {Cybern. Inf. Technol.}, 
month = sep, 
pages = {23–39}, 
numpages = {17}, 
}

@book{schur,
  title={The Schur complement and its applications},
  author={Zhang, Fuzhen},
  volume={4},
  year={2006},
  publisher={Springer Science \& Business Media}
}

@article{sbx,
    author = {Deb, Kalyanmoy and Ram Bhushan Agrawal},
    title = {Simulated binary crossover for continuous search space},
    journal = {Complex systems},
    year = {1995},
    volume = {9},
    number = {2},
    pages = {115-148}
}

@article{pm,
author = {Deb, Kalyanmoy and Deb, Debayan},
title = {Analysing mutation schemes for real-parameter genetic algorithms},
journal = {International Journal of Artificial Intelligence and Soft Computing},
volume = {4},
number = {1},
pages = {1-28},
year = {2014},
}

@misc{adam,
      title={Adam: A Method for Stochastic Optimization}, 
      author={Diederik P. Kingma and Jimmy Ba},
      year={2017},
      eprint={1412.6980},
      archivePrefix={arXiv},
      primaryClass={cs.LG},
      url={https://arxiv.org/abs/1412.6980}, 
}

@article{lhs,
author = {Wei-Liem Loh},
title = {{On Latin hypercube sampling}},
volume = {24},
journal = {The Annals of Statistics},
number = {5},
publisher = {Institute of Mathematical Statistics},
pages = {2058 -- 2080},
year = {1996},
}

@Article{nature,
author={Cully, Antoine
and Clune, Jeff
and Tarapore, Danesh
and Mouret, Jean-Baptiste},
title={Robots that can adapt like animals},
journal={Nature},
year={2015},
month={May},
day={01},
volume={521},
number={7553},
pages={503-507},
issn={1476-4687},
doi={10.1038/nature14422},
url={https://doi.org/10.1038/nature14422}
}

@ARTICLE{zhixing,
  author={Huang, Zhixing and Mei, Yi and Zhang, Fangfang and Zhang, Mengjie},
  journal={IEEE Trans. on Evol. Comput.}, 
  title={Multitask Linear Genetic Programming with Shared Individuals and its Application to Dynamic Job Shop Scheduling}, 
  year={2023},
  volume={},
  number={},
  pages={1-1},
  doi={10.1109/TEVC.2023.3263871}}

@ARTICLE{recommendation,
  author={Feng, Liang and Shang, Qingxia and Hou, Yaqing and Tan, Kay Chen and Ong, Yew-Soon},
  journal={IEEE Trans. on Artificial Intell.}, 
  title={Multispace Evolutionary Search for Large-Scale Optimization With Applications to Recommender Systems}, 
  year={2023},
  volume={4},
  number={1},
  pages={107-120},
  doi={10.1109/TAI.2022.3156952}}

@ARTICLE{wuyue,
  author={Wu, Yue and Ding, Hangqi and Gong, Maoguo and Qin, A. K. and Ma, Wenping and Miao, Qiguang and Tan, Kay Chen},
  journal={IEEE Trans. on Evol. Comput.}, 
  title={Evolutionary Multiform Optimization With Two-Stage Bidirectional Knowledge Transfer Strategy for Point Cloud Registration}, 
  year={2024},
  volume={28},
  number={1},
  pages={62-76},
  doi={10.1109/TEVC.2022.3215743}}

@ARTICLE{reconstruction1,
  author={Li, Hao and Ong, Yew-Soon and Gong, Maoguo and Wang, Zhenkun},
  journal={IEEE Trans. on Evol. Comput.},
  title={Evolutionary Multitasking Sparse Reconstruction: Framework and Case Study},
  year={2019},
  volume={23},
  number={5},
  pages={733-747},
  doi={10.1109/TEVC.2018.2881955}}

@ARTICLE{qiuxin,
  author={Qiu, Xin and Xu, Jian-Xin and Xu, Yinghao and Tan, Kay Chen},
  journal={IEEE Trans. on Cybern.}, 
  title={A New Differential Evolution Algorithm for Minimax Optimization in Robust Design}, 
  year={2018},
  volume={48},
  number={5},
  pages={1355-1368},
  doi={10.1109/TCYB.2017.2692963}}

@ARTICLE{momfea2,
  author={Bali, Kavitesh Kumar and Gupta, Abhishek and Ong, Yew-Soon and Tan, Puay Siew},
  journal={IEEE Trans. on Cybern.},
  title={Cognizant Multitasking in Multiobjective Multifactorial Evolution: {MO-MFEA-II}},
  year={2021},
  volume={51},
  number={4},
  pages={1784-1796}}

@misc{RAL-CBO,
      title={Controller Adaptation via Learning Solutions of Contextual {Bayesian} Optimization}, 
      author={Viet-Anh Le and Andreas A. Malikopoulos},
      year={2025},
      eprint={2403.04881},
      archivePrefix={arXiv},
      primaryClass={eess.SY},
      url={https://arxiv.org/abs/2403.04881}, 
}

@ARTICLE{xue-family,
  author={Xue, Xiaoming and Yang, Cuie and Feng, Liang and Zhang, Kai and Song, Linqi and Tan, Kay Chen},
  journal={IEEE Trans. on Cybern.}, 
  title={A Scalable Test Problem Generator for Sequential Transfer Optimization}, 
  year={2025},
  volume={55},
  number={5},
  pages={2110-2123},
  doi={10.1109/TCYB.2025.3547565}}

@article{bilevel,
  title={Evolutionary multitasking in bi-level optimization},
  author={Gupta, Abhishek and Ma{\'n}dziuk, Jacek and Ong, Yew-Soon},
  journal={Complex \& Intelligent Syst.},
  volume={1},
  number={1-4},
  pages={83--95},
  year={2015},
  publisher={Springer}
}

\clearpage
\onecolumn












%
\title{Supplementary File of ``($\boldsymbol{\theta}_l, \boldsymbol{\theta}_u$)-Parametric Multi-Task Optimization: Joint Search in Solution and Infinite Task Spaces''}
\maketitle

\setcounter{section}{0}
\renewcommand\thesection{S-\Roman{section}}

\setcounter{table}{0}
\renewcommand\thetable{S-\Roman{table}}

\setcounter{figure}{0}
\renewcommand\thefigure{S-\arabic{figure}}

\setcounter{equation}{0}
\renewcommand\theequation{S-\arabic{equation}}


\section{Synthetic Problems}
The synthetic parametric multi-task optimization problems are constructed based on canonical continuous benchmark problems, modified to include task parameters.
The objective function is defined as:
\begin{equation}
     f(\mathbf{x}, \boldsymbol{\theta}) = g(\lambda(\mathbf{x}-\boldsymbol{\sigma}(L\boldsymbol{\theta}))), \mathbf{x}\in [0,1]^{N}, \boldsymbol{\theta} \in [0,1]^{D}
     \label{eq:skeleton}
\end{equation}
where $g(\sbullet)$ is the base objective function, such as continuous optimization functions including Sphere, Ackley, Rastrigin, and Griewank to model the optimization problem, $\mathbf{x}$ represents the decision variables within the $N$-dimensional solution space, $\boldsymbol{\theta}$ denotes the task parameter within the $D$-dimensional task space, $\lambda > 0$ is a scaling factor to adjust the magnitude of the decision variable, $L \in \mathbb{R}^{N \times D}$ is a linear transformation matrix that maps the task parameter into the $N$-dimensional solution space, and $\boldsymbol{\sigma}$ represents a nonlinear transformation applied to the transformed task parameter.
One can refer to TABLE \ref{tab: synthetic} for details of the synthetic problems in our paper.
The source code of this paper can be found at: https://github.com/ambigeV/pmto-tevc.
\begin{table}[htbp]
\centering
\caption{\textcolor{black}{Details of Synthetic Problems}}
\label{tab: synthetic}
\begin{tabular}{c|c|c|c|c}
\hline
Problems     & $g(\sbullet)$ & $L$ & $\lambda$ & $\boldsymbol{\sigma}(\sbullet)$ \\ \hline\hline
Sphere-I     & Sphere        & $L$                     & 4                        & $\boldsymbol{\sigma}_1(\sbullet)$            \\ \hline
Sphere-II    & Sphere        & $L$                     & 4                        & $\boldsymbol{\sigma}_2(\sbullet)$            \\ \hline\hline
Ackley-I     & Ackley        & $L$                     & 4                        & $\boldsymbol{\sigma}_1(\sbullet)$            \\ \hline
Ackley-II    & Ackley        & $L$                     & 4                        & $\boldsymbol{\sigma}_2(\sbullet)$            \\ \hline\hline
Rastrigin-I  & Rastrigin     & $L$                     & 20                        & $\boldsymbol{\sigma}_1(\sbullet)$            \\ \hline
Rastrigin-II & Rastrigin     & $L$                     & 20                        & $\boldsymbol{\sigma}_2(\sbullet)$            \\ \hline\hline
Griewank-I   & Griewank      & $L$                     & 600                        & $\boldsymbol{\sigma}_1(\sbullet)$            \\ \hline
Griewank-II  & Griewank      & $L$                    & 600                        & $\boldsymbol{\sigma}_2(\sbullet)$            \\ \hline\hline
\end{tabular}
\end{table}

In the TABLE \ref{tab: synthetic} the linear transformation adopted in this paper is as follows (we assume the task space has 5 dimensions and the solution space has 4 dimensions):
\begin{equation}
  L = 
\begin{bmatrix}
1 & 0 & 0 & 0 & 0 \\
0 & 2/3 & 1/3 & 0 & 0 \\
0 & 0 & 1/3 & 2/3 & 0 \\
0 & 0 & 0 & 0 & 1 \\
\end{bmatrix}
\end{equation}
and two nonlinear transformations can be defined as follows:
\begin{equation}
    \boldsymbol{\sigma}_1(x) = (\sin(5 * (x + 0.5)) + 1) / 2
\end{equation}
\begin{equation}
    \boldsymbol{\sigma}_2(x) = 0.3 * (1 + \sin(5 \pi x - \pi / 2)) + 0.3(x - 0.2)^2
\end{equation}
where the first nonlinear mapping has lower frequency of nonlinear component compared to the second nonlinear mapping, indicating an easier optimization process.
Also, both nonlinear mappings ensure the solutions can be mapped from $[0,1]^N$ to $[0,1]^N$.
And finally, we recap the canonical formulations of the adopted continuous benchmark functions as:
\begin{itemize}
    \item Sphere function: $g(\mathbf{z}) = \sum_{i=1}^N z_i^2$
    \item Ackley function: $g(\mathbf{z})=20+e-20\exp(-0.2\sqrt{\frac{1}{N}\sum_{i=1}^N z_i^2})-\exp(\frac{1}{N}\sum_{i=1}^N \cos(2\pi z_i))$
    \item Rastrigin function: $g(\mathbf{z}) = \sum_{i=1}^N (z_i^2-10\cos(2\pi z_i)+10)$
    \item Griewank function: $g(\mathbf{z}) = \sum_{i=1}^N z_i^2/4000 - \Pi_{i=1}^N \cos(z_i/\sqrt{i})+1$
\end{itemize}

\section{Crane-Load System Optimization}
The Crane-Load system optimization problem is to control the duration so that an external force can be switched from an upper constraint to the lower one, so that the system can be accelerated from a rest state to a steady goal velocity with minimal time and oscillation~\cite{crane}.
The decision variables of Crane-Load system optimization problems are the time intervals to decide how long a specific drive force $F$ is exerted on the system as shown in Fig.~3(b) in the manuscript.
The goal is to optimize an aggregated function to minimize the overall control time $t_1 + t_2 + t_3$ and achieve the desired velocity for the system with minimal oscillation.
The total objective function of this problem can be illustrated as follows:
\begin{equation}
    \argmin_{t_1,t_2,t_3} \frac{2E}{m_2v^2}+(t_1+t_2+t_3)\frac{\Omega}{2\pi},
    \label{eq:goal}
\end{equation}
\begin{equation}
    E = 
    \begin{cases}
    \begin{aligned}
        w \cdot TE, TE \geq \Delta\\
        0, TE < \Delta
    \end{aligned}
    \end{cases}
\end{equation}
where the first term in \ref{eq:goal} indicates the specific energy of the whole system and the second term is the system acceleration duration. 
$TE$ denotes the terminal energy of the system, reflecting the level of oscillation as follows:
\begin{equation}
\begin{aligned}
    TE = &\frac{m_2}{(2m_1^2\Omega^6)}
             (\Omega^2\Omega_0^4)
              (F_{max} - W - (F_{max} - F_{min})
               (\cos(t_3\Omega) - \cos((t_2 + t_3)\Omega)) +
              (W - F_{max})\cos((\sum_{i=1}^3(t_i) \Omega)^2) + \\
              &\Omega_4((F_{max} - F_{min})*(\sin(t_3\Omega) - \sin((t_2 + t_3)\Omega))) +
              (F_{max} - W) \sin((\sum_{i=1}^3(t_i)\Omega)^2) + TE_2^2
              ),
\end{aligned}
\end{equation}
\begin{equation}
\begin{aligned}
TE_2 = &(m_1 v \Omega^3 - \Omega \Omega_0^2 
         (F_{min} * t_2 + F_{max} * (t_1 + t_3) - \sum_{i=1}^3(t_i) W) +\\
         &\Omega_0^2 * ((F_{max} - F_{min})*(\sin(t_3\Omega) - \sin((t_2 + t_3)\Omega)) +
         (F_{max} - W) * \sin(t\sum_{i=1}^3(t_i) \Omega))).
\end{aligned}
\end{equation}
\begin{table}[]
\centering
\caption{\textcolor{black}{Parameter Details of Crane-Load System-I Optimization Problems}}
\label{tab: crane-I}
\begin{tabular}{|c|c|}
\hline
Variable & Value    \\ \hline
$m_2$     & 1.00E+04 \\ \hline
$m_1$     & 4.20E+04 \\ \hline
$v$        & 0.7      \\ \hline
$l$        & 6.5      \\ \hline
$W$        & $0.01g(m_1+m_2)$     \\ \hline
$w$        & 1.00E+06 \\ \hline
$\Delta$    & 0.01     \\ \hline
$F_{min}$   & 0        \\ \hline
$F_{max}$   & 2.41E+04 \\ \hline
$\Omega$    & $\sqrt{g(m_1+m_2)/m_1l}$ \\ \hline
$\Omega_0$    & $\sqrt{g/l}$ \\ \hline
$t_i$ & $[0,2]$ \\ \hline
$\Delta t_i$ & $[0,1]$ \\ \hline
\end{tabular}
\end{table}
\begin{table}[]
\centering
\caption{\textcolor{black}{Parameter Details of Crane-Load System-II Optimization Problems}}
\label{tab: crane-II}
\begin{tabular}{|c|c|}
\hline
Variable & Value    \\ \hline
$m_{2,{\min}}$     & 0.80E+03 \\ \hline
$m_{2,{\max}}$     & 1.20E+04 \\ \hline
$m_1$     & 4.20E+04 \\ \hline
$v$        & 0.7      \\ \hline
$l_{\min}$        & 5      \\ \hline
$l_{\max}$        & 8      \\ \hline
$W_{\min}$        & $0.005g(m_1+m_2)$     \\ \hline
$W_{\max}$        & $0.015g(m_1+m_2)$     \\ \hline
$w$        & 1.00E+06 \\ \hline
$\Delta$    & 0.01     \\ \hline
$F_{min}$   & 0        \\ \hline
$F_{max}$   & 2.41E+04 \\ \hline
$\Omega$    & $\sqrt{g(m_1+m_2)/m_1l}$ \\ \hline
$\Omega_0$    & $\sqrt{g/l}$ \\ \hline
$t_i$ & $[0,3]$ \\ \hline
\end{tabular}
\end{table}

For the Crane-Load System-I Optimization, we consider the possible time delays for each control variable $t_1, t_2, t_3$, so that the task parameters are $\Delta t_1, \Delta t_2, \Delta t_3$ and when evaluating the original objective function we substitute the control variable via $t_1+\Delta t_1, t_2+\Delta t_2, t_3+\Delta t_3$ and parameters in TABLE \ref{tab: crane-I}.
For the Crane-Load System-II Optimization, we consider the distinct loading cases for parameter $m_2, l, W$ as task parameters.
When evaluating the original objective function we substitute the actual environmental parameter $m_2, l, W$ by the actual task parameter within the ranges $(m_{2,{\min}}, m_{2,{\max}}), (l_{\min}, l_{\max}), (W_{\min}, W_{\max})$, and evaluate the control variables $t_1, t_2, t_3$ under these environmental parameters.
One can refer to the specific parameters in TABLE \ref{tab: crane-II} and more details in \cite{crane}.

\section{Truss Design Problems}
The truss design problem focuses on optimizing structural efficiency by minimizing the structural weight (volume) while controlling joint stress, a common benchmark in multiobjective optimization literature~\cite{real-bench}. The analysis assumes the truss is positioned with fixed horizontal dimensions, and the structural design is governed by three independent variables $\boldsymbol{\theta}=(\theta_1, \theta_2, \theta_3)$.

The base objective functions, which are aggregated for the single-objective PMTO framework, are:
\begin{equation}
    \begin{cases}
    \begin{aligned}
        & f_1 = \theta_1\sqrt{16+\theta_3^2}+\theta_2\sqrt{1+\theta_3^2}\\
        & f_2 = \frac{20\sqrt{16+\theta_3^2}}{\theta_1\theta_3}
    \end{aligned}
    \end{cases}
    \label{eq:truss}
\end{equation}
The variables $\theta_i$ represent the following physical properties (Task Parameters):
\begin{itemize}
    \item $\theta_1$: Cross-sectional \textbf{Area} of Bar 1.
    \item $\theta_2$: Cross-sectional \textbf{Area} of Bar 2.
    \item $\theta_3$: \textbf{Vertical Distance} (Height) of the central loaded joint.
\end{itemize}
The total aggregated objective is $f = \alpha_1 f_1 + \alpha_2 f_2$, where $\alpha_1=10.00$ and $\alpha_2=1.00\text{E-5}$.
The \textbf{operating structural parameters} used in the objective function $f$ are defined by the nominal design plus the processing error: $\boldsymbol{\theta} + \mathbf{x}$, where each $x_i$ corresponds to the measurement error of physical property $\theta_i$.
As mentioned in the main text, to objective this aggregated objective function $f(\mathbf{x}, \boldsymbol{\theta})$ in a minimax flavor, we treat $\mathbf{x}$ as decision variables and $\boldsymbol{\theta}$ as task parameters.
\begin{itemize}
    \item \textbf{Task Parameters ($\boldsymbol{\theta}$):} These are the nominal design values we seek to optimize:
        $$\boldsymbol{\theta} \in [\theta_{1}, \theta_{2}, \theta_{3}] \quad \text{where} \quad \theta_1, \theta_2 \in [2, 100] \quad \text{and} \quad \theta_3 \in [1, 3]$$

    \item \textbf{Decision Variables ($\mathbf{x}$):} This vector $\mathbf{x}=(x_1, x_2, x_3)$ is the input to the inner maximization, representing the \textbf{fractional processing error} for each measurement:
    $$x_i = x_{i,frac} \cdot (\theta_{i,max}-\theta_{i,min}),$$ 
        $$x_{i,frac} \in [-0.05, 0.05] \quad \text{for all} \quad i=1, 2, 3$$
    This is modeled as a fractional deviation on the operating parameter relative to its full search range.
\end{itemize}

\section{Proof of the Theorem}
\begin{theorem}
    Let the kernel functions used in PMTO-FT and the independent strategy satisfy $\kappa_{pmt}((\mathbf{x}, \boldsymbol{\theta}),(\mathbf{x}', \boldsymbol{\theta}))=\kappa_{ind}(\mathbf{x},\mathbf{x}')$. Then, the MIG in the independent strategy, denoted as $\gamma_{T,m}^{ind}$, and the MIG in the PMTO-FT, denoted as $\gamma_{T,m}$, satisfy $\gamma_{T,m} \leq \gamma_{T,m}^{ind}, \forall m \in [M], \forall T \geq 1$.
\end{theorem}
\begin{IEEEproof}
Given the dataset $\mathcal{D}_{pmt}$, in the multivariate Gaussian case, the \textit{conditional information gain} about the objective function corresponding to arbitrary parameterized task of interest $m^* \in [M]$ is:
\begin{equation}    I([y^{(1)},\ldots,y^{(T)}];\textit{\textbf{f}}_{m^{*}}|\mathcal{D}_{pmt}) = \frac{1}{2}\log|\mathbf{I}+\sigma_{\epsilon}^{-2}\mathbf{K}_{pmt, m^{*}}|,
\end{equation}
where $\mathbf{K}_{pmt, m^{*}}$ denotes the conditional covariance matrix for the parameterized task $m^{*}$, derived from the dataset $\mathcal{D}_{pmt}$, which includes data from all tasks.
Next, we show how to calculate $\mathbf{K}_{pmt, m^{*}}$.
Let $\pi_i, i\in[M]$ represents a rearrangement of the values $\{1, 2, \ldots, M\}$, then $\mathbf{K}_{pmt}$ can be further denoted as follows:
\begin{equation}
\mathbf{K}_{pmt} = 
\begin{bmatrix}
\mathbf{K}_{\pi_1,\pi_1} & \mathbf{K}_{\pi_1,\pi_2} & \cdots & \mathbf{K}_{\pi_1,\pi_M} \\
\mathbf{K}_{\pi_2,\pi_1} & \mathbf{K}_{\pi_2,\pi_2} & \cdots & \mathbf{K}_{\pi_2,\pi_M} \\
\vdots & \vdots & \ddots & \vdots \\
\mathbf{K}_{\pi_M,\pi_1} & \mathbf{K}_{\pi_M,\pi_2} & \cdots & \mathbf{K}_{\pi_M,\pi_M}
\end{bmatrix},
\label{eq:kernel_matrix}
\end{equation}
where each block matrix $\mathbf{K}_{i,j} \in \mathbb{R}^{T \times T}$ can be defined as follows:
\begin{equation}
\begin{aligned}
&\mathbf{K}_{i,j} = \\
&\begin{bmatrix}
\kappa((\mathbf{x}_{1,i}, \boldsymbol{\theta}_i),(\mathbf{x}_{1,j}, \boldsymbol{\theta}_j)) & \cdots & \kappa((\mathbf{x}_{1,i}, \boldsymbol{\theta}_i),(\mathbf{x}_{T,j}, \boldsymbol{\theta}_j)) \\
\vdots & \ddots & \vdots \\
\kappa((\mathbf{x}_{T,i}, \boldsymbol{\theta}_i),(\mathbf{x}_{1,j}, \boldsymbol{\theta}_j)) & \cdots & \kappa((\mathbf{x}_{T,i}, \boldsymbol{\theta}_i),(\mathbf{x}_{T,j}, \boldsymbol{\theta}_j))
\end{bmatrix}.
\end{aligned}
\label{eq:kernel_function}
\end{equation}
Without loss of generality, let $\pi_{M} = m^{*}$, 
\begin{equation}
\mathbf{K}_{\setminus m^{*}} = 
\begin{bmatrix}
\mathbf{K}_{\pi_1,\pi_1} & \cdots & \mathbf{K}_{\pi_1,\pi_{M-1}} \\
\vdots & \ddots & \vdots \\
\mathbf{K}_{\pi_{M-1},\pi_1} & \cdots & \mathbf{K}_{\pi_{M-1},\pi_{M-1}}
\end{bmatrix},
\end{equation}
and $B = \begin{bmatrix} \mathbf{K}_{\pi_M,\pi_1} & \mathbf{K}_{\pi_M,\pi_2} & \cdots & \mathbf{K}_{\pi_M,\pi_M} \end{bmatrix}^T$, then we can further formulate $\mathbf{K}_{pmt}$ as:
\begin{equation}
\mathbf{K}_{pmt} = 
\begin{bmatrix}
\mathbf{K}_{\setminus m^{*}} & B \\
B^T & \mathbf{K}_{m^*,m^*}
\end{bmatrix}.
\label{eq:simplified_k_matrix}
\end{equation}
Using the properties of the conditional Gaussian distribution, $\mathbf{K}_{pmt, m^{*}}$ can be calculated as:
\begin{equation}
\begin{aligned}
\mathbf{K}_{pmt, m^{*}} = \mathbf{K}_{m^*,m^*} - B^T(\mathbf{K}_{\setminus m^{*}} + \sigma_{\epsilon}^{-2}\mathbf{I})^{-1}B.
\end{aligned}
\label{eq:schur}
\end{equation}
Likewise, in terms of the independent strategy, the corresponding conditional information gain is
\begin{equation}
    I([y^{(1)},\ldots,y^{(T)}];\textit{\textbf{f}}_{m^{*}}|\mathcal{D}_{m^{*}}) = \frac{1}{2}\log|\mathbf{I}+\sigma_{\epsilon}^{-2}\mathbf{K}_{ind,m^{*}}|,
\end{equation}
where $\mathbf{K}_{ind,m^{*}}$ is the covariance matrix with all the samples in $\mathcal{D}_{m^{*}}$. Based on the assumption on the kernel function that $\kappa_{pmt}((\mathbf{x}, \boldsymbol{\theta}_m),(\mathbf{x}', \boldsymbol{\theta}_m)) = \kappa_{ind}(\mathbf{x}, \mathbf{x}')$, it follows that $\mathbf{K}_{ind,m^{*}} = \mathbf{K}_{m^*,m^*}$. This can further help us to analyze the relationship of the two conditional information gains. Since the covariance matrix $\mathbf{K}_{pmt}$, $\mathbf{K}_{\setminus m^{*}}$ and $\mathbf{K}_{m^{*}, m^{*}}$ are both positive semi-definite (PSD), then according to the \textit{Schur complement theorem}~\cite{schur}, we know $\mathbf{K}_{m^*,m^*} - B^T(\mathbf{K}_{\setminus {m^*}} + \sigma_{\epsilon}^{-2}\mathbf{I})^{-1}B$ and $\mathbf{K}_{\setminus {m^*}} + \sigma_{\epsilon}^{-2}\mathbf{I}$ are also PSD. 
Then, considering the \textit{Minkowski determinant inequality} that for PSD matrices $C$ and $D$, we have $|C + D| \geq |C|+|D| \geq |C|$. Substituting $C$ and $D$ by $C = \mathbf{I}+\sigma_{\epsilon}^{-2}{\mathbf{K}}_{pmt,m^{*}}$ and $D = \mathbf{I}+\sigma_{\epsilon}^{-2}B^T(\mathbf{K}_{\setminus m^{*}} + \sigma_{\epsilon}^{-2}\mathbf{I})^{-1}B$, respectively, then the follows can be obtained:
\begin{equation}
|\mathbf{I} + \sigma_{\epsilon}^{-2} \mathbf{K}_{m^{*}, m^{*}}| = |\mathbf{I} + \sigma_{\epsilon}^{-2} \mathbf{K}_{ind, m^{*}}| \geq |\mathbf{I}+\sigma_{\epsilon}^{-2} {\mathbf{K}}_{pmt, m^{*}}|.
\end{equation}
This results in the fact that 
\begin{equation}
I([y^{(1)},\ldots,y^{(T)}]; \textit{\textbf{f}}_{m^{*}}|\mathcal{D}_{m^{*}}) \geq I([y^{(1)},\ldots,y^{(T)}];\textit{\textbf{f}}_{m^{*}}|\mathcal{D}_{pmt}).
\label{eqn:relationship_I}
\end{equation}
Hence, we can deduce from \eqref{eqn:relationship_I} that $\gamma_{T,m^{*}} \leq \gamma_{T,m^{*}}^{ind}$.
\end{IEEEproof}

\section{Sensitivity Analysis}

\begin{figure}
\centering
\includegraphics[width=6in]{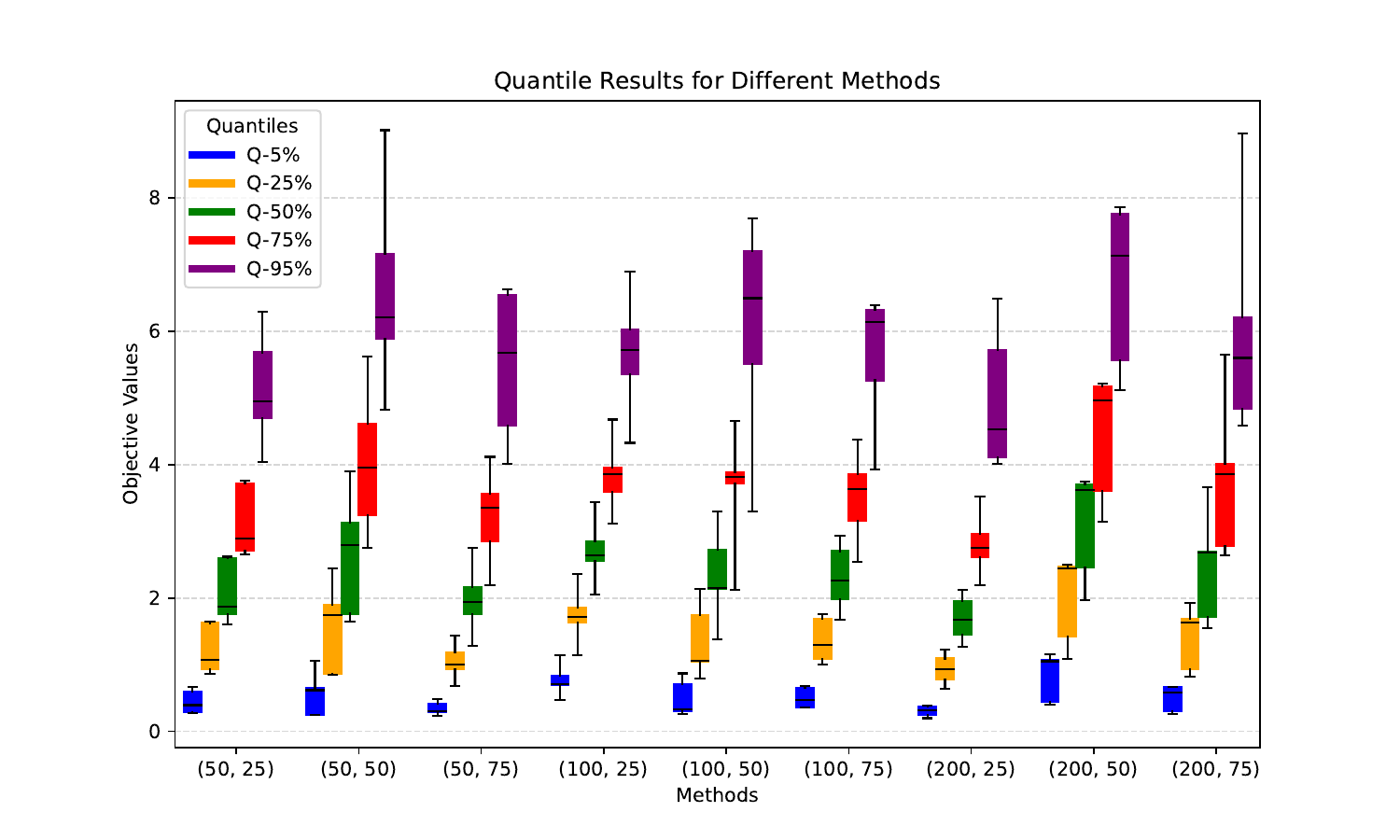}%
\caption{Comparison of the task model's online performance under varying processing errors for the proposed PMTO method on Sphere-II. Distinct tuple values indicate the choices of ($P$, $G$), where $P$ is the population size and $G$ is the generation size of the adopted EA optimizer.}
\label{fig: sphere-ii}
\end{figure}

\begin{figure}
\centering
\includegraphics[width=6in]{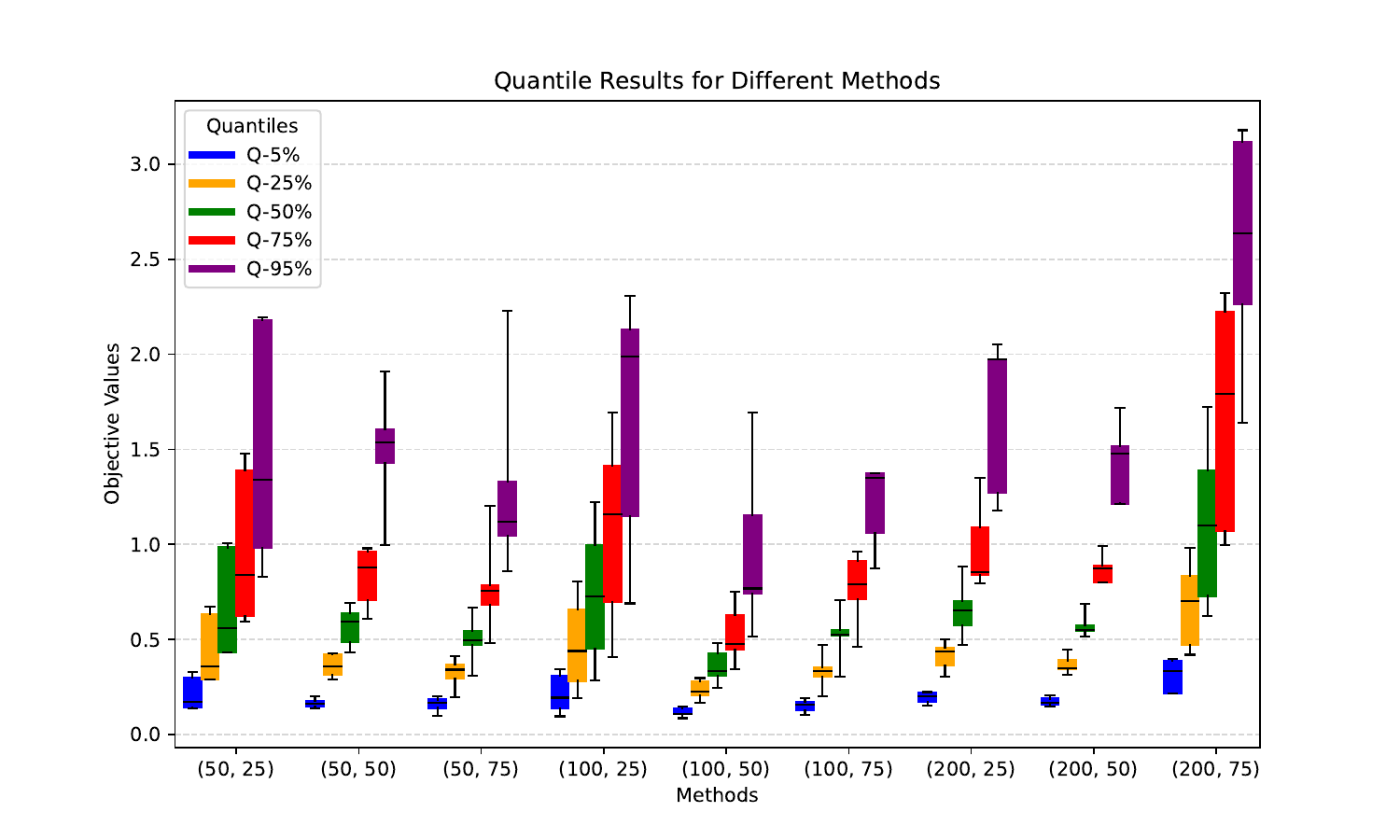}%
\caption{Comparison of the task model's online performance under varying processing errors for the proposed PMTO method on Ackley-I. Distinct tuple values indicate the choices of ($P$, $G$), where $P$ is the population size and $G$ is the generation size of the adopted EA optimizer.}
\label{fig: ackley-i}
\end{figure}

Since other hyperparameters of the proposed method more or less align with the well-established default settings, herein we only study the influence of two hyperparameters in the task evolution module.
To assess the robustness of our method with respect to the choice of hyperparameters in the task evolution module, we conduct a sensitivity analysis on the population size $P$ and the number of generations $G$. 
Specifically, we vary $P \in \{50, 100, 150\}$ and $G \in \{25, 50, 75\}$, and evaluate the task model's online optimization performance under each configuration. 
As summarized in Fig.~\ref{fig: sphere-ii} and Fig.~\ref{fig: ackley-i}, the proposed method maintains relatively consistent performance across a wide range of $(P, G)$ settings. These results indicate that the method is not overly sensitive to the choice of $P$ and $G$, and that reliable performance can be achieved without extensive hyperparameter tuning.

\onecolumn
\clearpage
\end{document}